\documentclass[10pt,twocolumn,letterpaper]{article}

\usepackage{times}
\usepackage{microtype}
\usepackage{epsfig}
\usepackage[table,xcdraw]{xcolor}
\usepackage{caption}
\usepackage{float}
\usepackage{placeins}
\usepackage{color, colortbl}
\usepackage{stfloats}
\usepackage{enumitem}
\usepackage{tabularx}
\usepackage{xstring}
\usepackage{multirow}
\usepackage{xspace}
\usepackage{url}
\usepackage{subcaption}
\usepackage{xcolor}
\usepackage[hang,flushmargin]{footmisc}
\usepackage{cuted}

\definecolor{synapse1}{RGB}{0,255,51}
\definecolor{synapse2}{RGB}{0,255,255}
\definecolor{synapse3}{RGB}{175,175,175}
\definecolor{synapse4}{RGB}{255,0,255}
\definecolor{synapse5}{RGB}{255,149,16}

% ...

\newcommand{\R}[1]{{%
    \textbf{%
        \ifstrequal{#1}{1}{\textcolor{red}{R#1}}{%
        \ifstrequal{#1}{2}{\textcolor{blue}{R#1}}{%
        \ifstrequal{#1}{3}{\textcolor{magenta}{R#1}}{%
        \ifstrequal{#1}{4}{\textcolor{teal}{R#1}}{%
                           \textcolor{cyan}{R#1}%
        }}}}%
    }%
}}

\usepackage{iccv}
\usepackage{times}
\usepackage{epsfig}
\usepackage{graphicx}
\usepackage{amsmath}
\usepackage{amssymb}
\usepackage{booktabs}

\usepackage[pagebackref=true,breaklinks=true,letterpaper=true,colorlinks,bookmarks=false]{hyperref}

\iccvfinalcopy 

\newcommand{\ehsan}[1]{\textcolor{red}{Ehsan: #1}}
\newcommand{\Yousef}[1]{\textcolor{purple}{Yousef: #1}}

\def\iccvPaperID{3961} 

\usepackage{tikz}
\usetikzlibrary{pgfplots.groupplots,matrix,positioning,calc,decorations.pathreplacing,arrows.meta}
\usepackage[autostyle]{csquotes}
\newcommand*\circled[1]{\tikz[baseline=(char.base)]{
            \node[shape=circle,draw,inner sep=1pt] (char) {#1};}}

\usepackage[usestackEOL]{stackengine}

\usepackage{tabularx}
\usepackage{multicol}

\usepackage{multirow}

\definecolor{ConvColorLegend}{RGB}{255,231,191}
\definecolor{ConvReluColorLegend}{RGB}{255,188,125}
\definecolor{PoolColorLegend}{RGB}{233,131,128}
\definecolor{SoftmaxColorLegend}{RGB}{157,88,159}
\definecolor{ConcatColorLegend}{RGB}{159,191,161}
\definecolor{UnpoolColorLegend}{RGB}{95,138,185}

\newcommand{\ahmad}[1]{\textcolor{blue}{#1}}
\newcommand{\Ahmad}[1]{\textcolor{cyan}{Ahmad: #1}}
\def\methodName{DIAMANT}
\usepackage[capitalize]{cleveref}
\crefname{section}{Sec.}{Secs.}
\crefname{table}{Table}{Tables}
\crefname{figure}{Fig.}{Figs.}

\ificcvfinal\fi

\begin{document}

\title{\methodName{}\\Dual Image-Attention Map Encoders For Medical Image Segmentation}

\author{Yousef Yeganeh${}^{1,3}$ \thanks{Equal contribution} \hspace{0.9cm} 
Azade Farshad${}^{1,3}$ \footnotemark[1]  \hspace{0.9cm}
Peter Weinberger${}^{1}$ \footnotemark[1]  \hspace{0.9cm} 
Seyed-Ahmad Ahmadi${}^{4}$   \\ \vspace{-0.25cm} \\ 
Ehsan Adeli${}^{5}$ \hspace{0.9cm} 
Nassir Navab${}^{1,2}$ 
\and
${}^{1}$Technical University of Munich
\and
${}^{2}$Johns Hopkins University
\and
${}^{3}$MCML
\and 
${}^{4}$NVIDIA
\and
${}^{5}$Stanford
}
\maketitle
\ificcvfinal\thispagestyle{empty}\fi

\begin{strip}
\begin{center}
\centering
\vspace{-1.5cm}
\resizebox{\textwidth}{!}{
 \begin{tabular}{c@{\hskip 0.15cm}c@{\hskip 0.15cm}c@{\hskip 0.15cm}c@{\hskip 0.15cm}c@{\hskip 0.15cm}c@{\hskip 0.15cm}c@{\hskip 0.15cm}c@{\hskip 0.15cm}c@{\hskip 0.15cm}c@{\hskip 0.15cm}}
& \textbf{Input} & \textbf{Attention Head 1} & \textbf{Attention Head 2} & \textbf{Attention Head 3} & \textbf{Attention Head 4} & \textbf{Attention Head 5} & \textbf{Attention Head 6} & \textbf{Segmentation} \\ 
{\textbf{A}} & 
\raisebox{-.5\height}{\includegraphics[width=3cm]{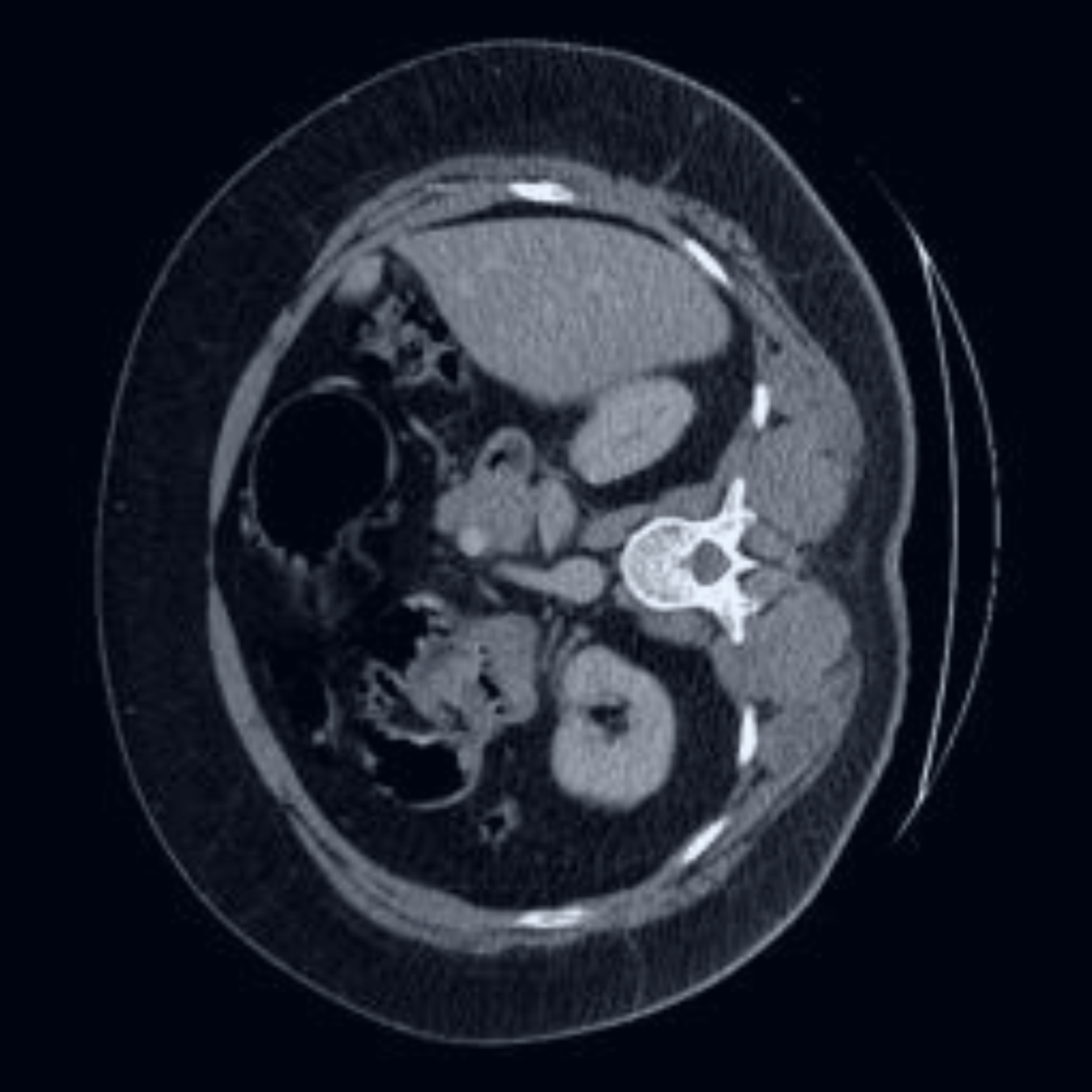}} &
 \raisebox{-.5\height}{\includegraphics[width=3cm]{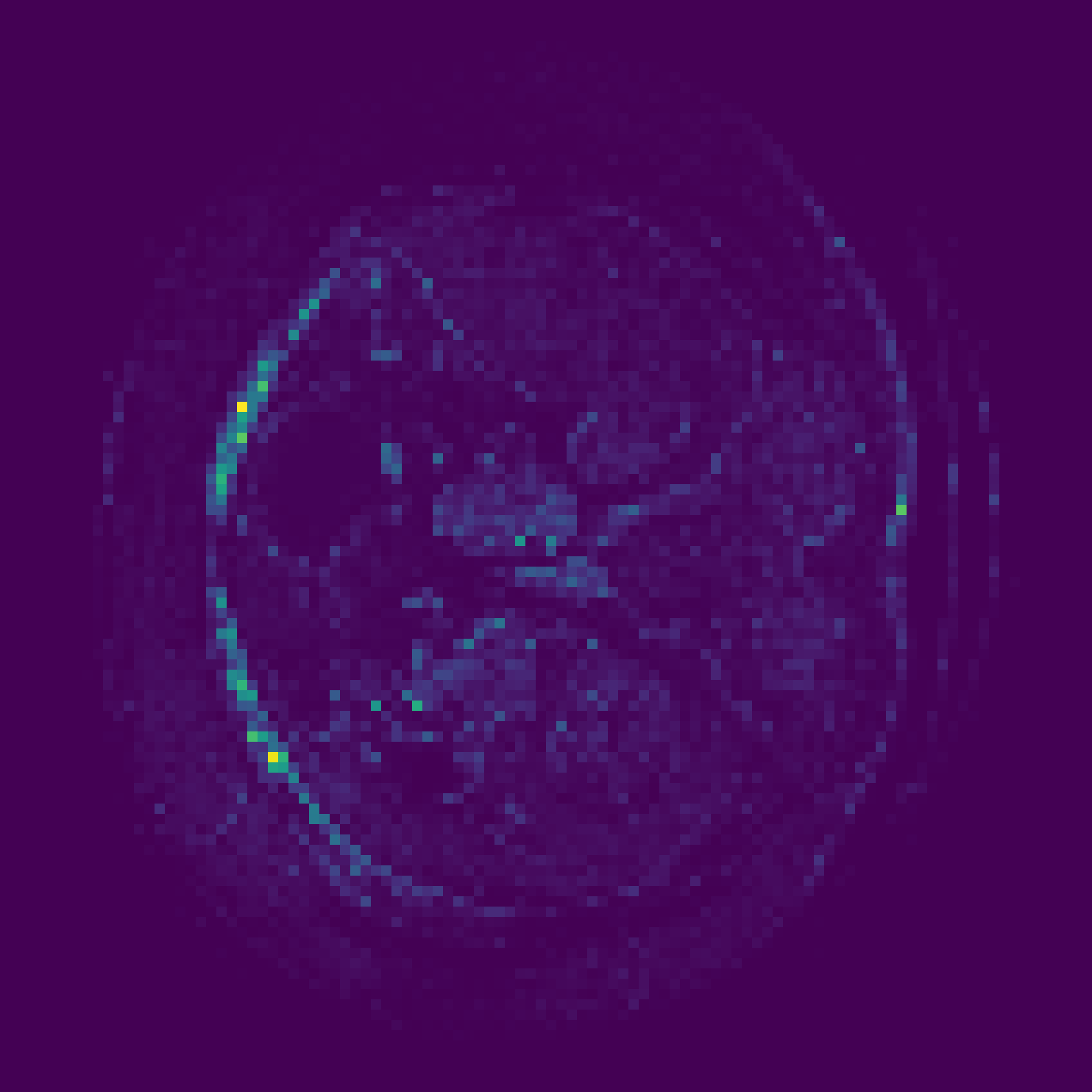}} &
  \raisebox{-.5\height}{\includegraphics[width=3cm]{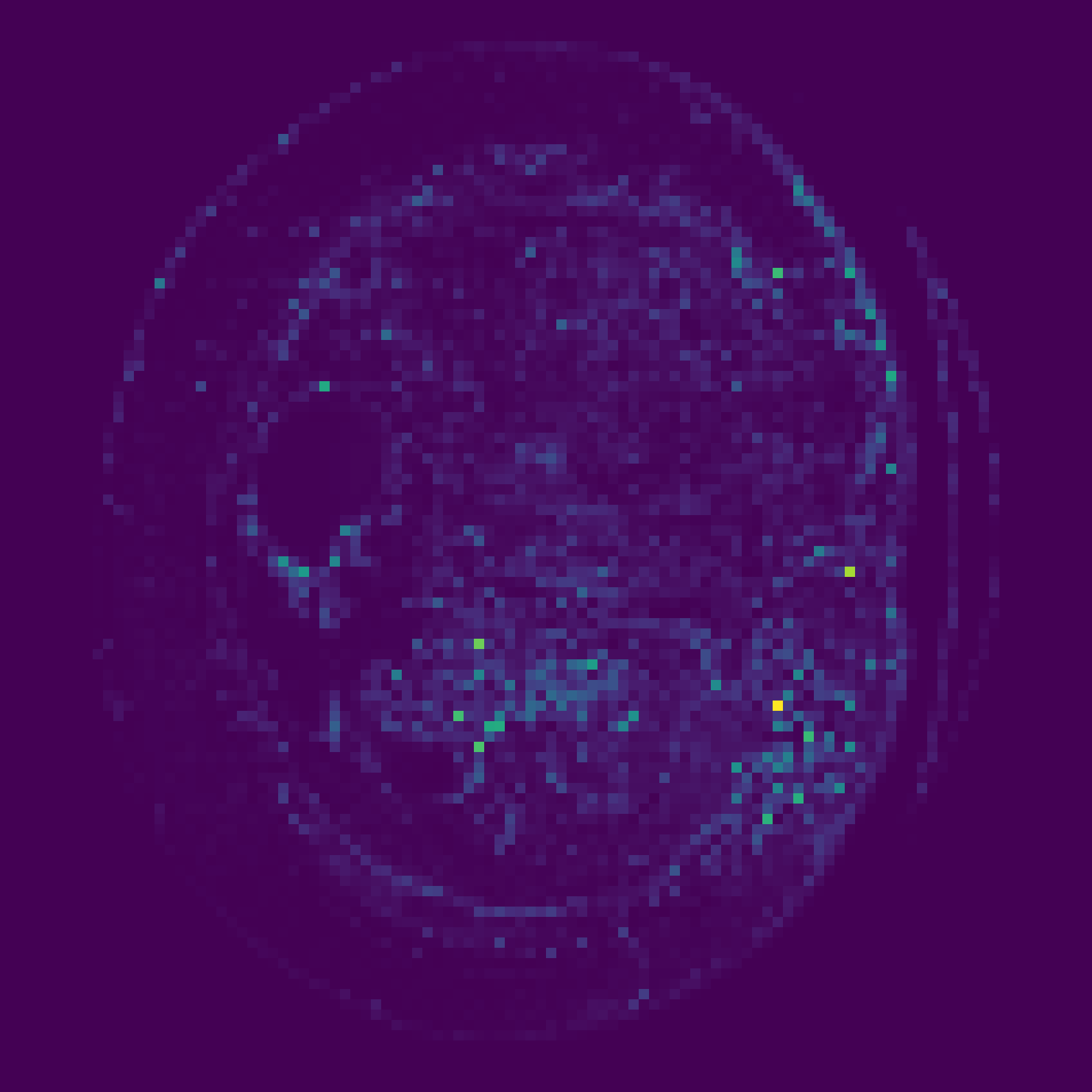}} &
   \raisebox{-.5\height}{\includegraphics[width=3cm]{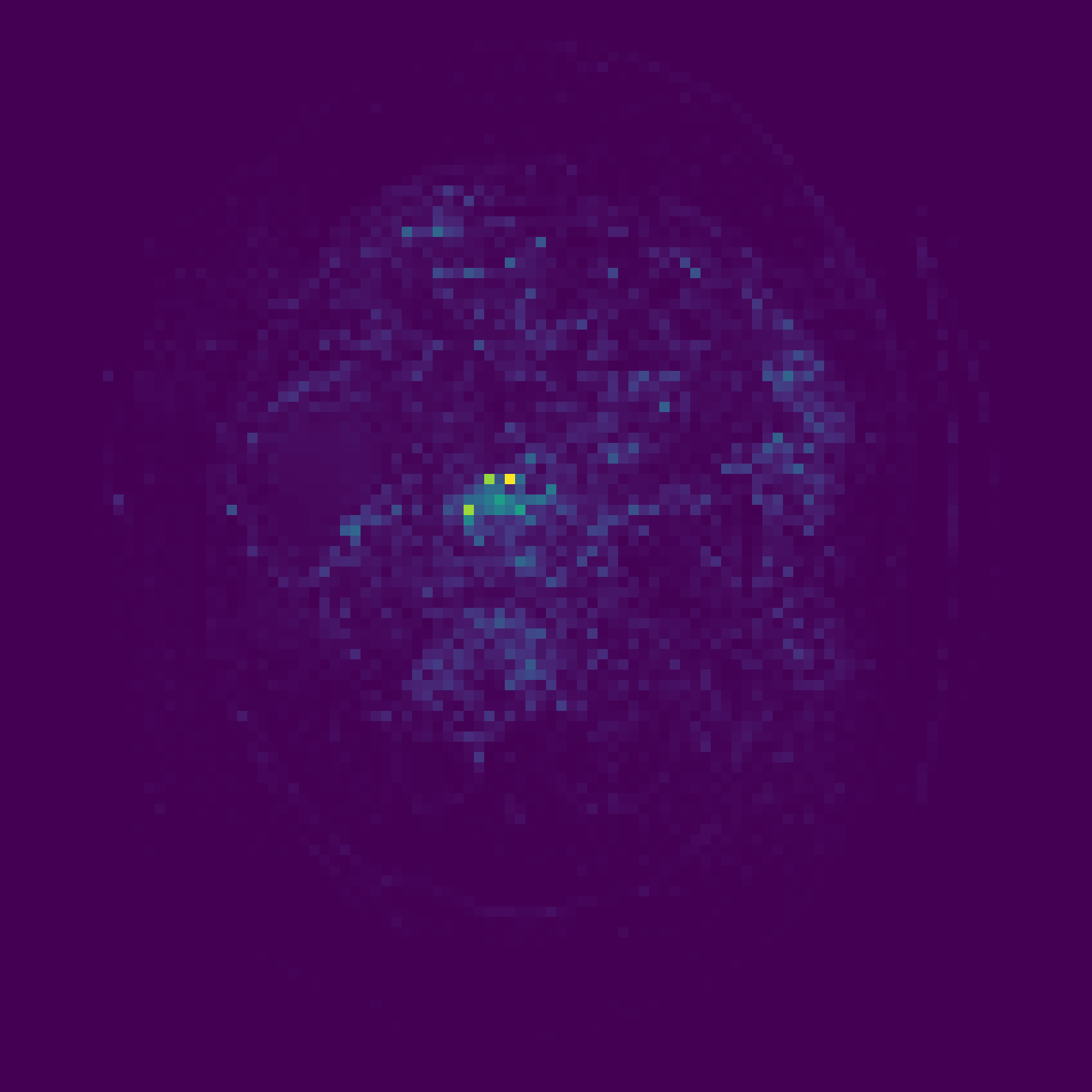}} &
    \raisebox{-.5\height}{\includegraphics[width=3cm]{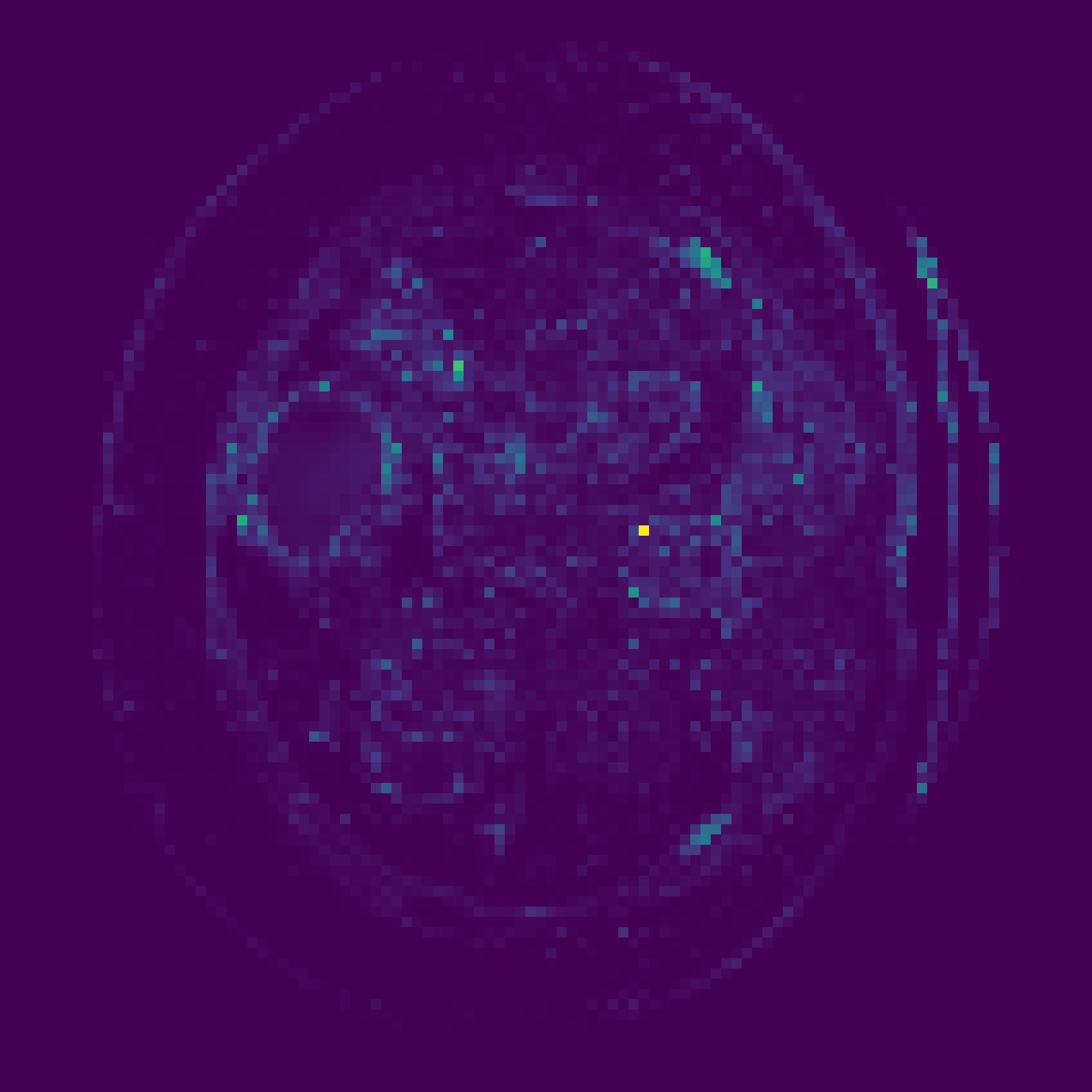}} &
     \raisebox{-.5\height}{\includegraphics[width=3cm]{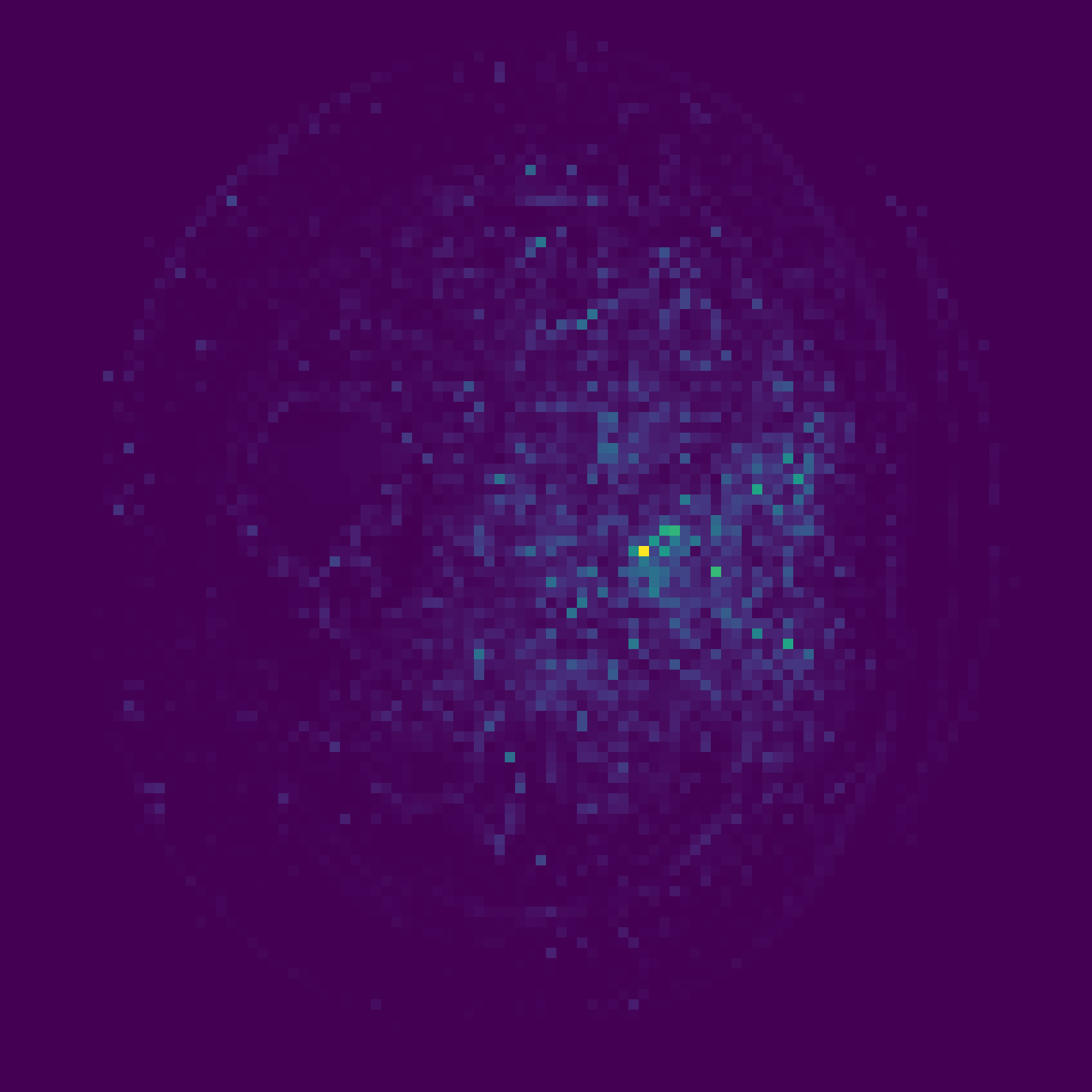}} &
      \raisebox{-.5\height}{\includegraphics[width=3cm]{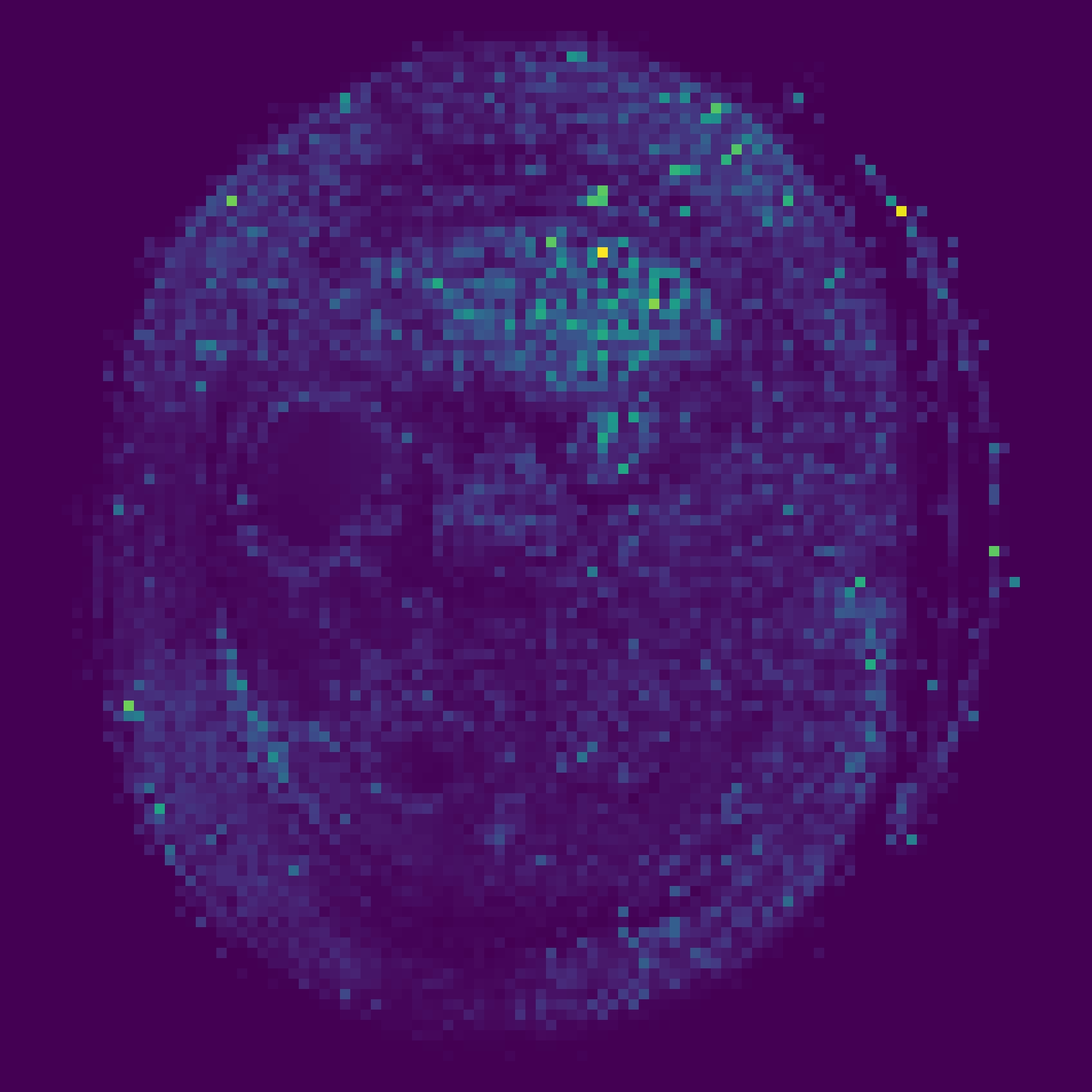}} &
 \raisebox{-.5\height}{\includegraphics[width=3cm]{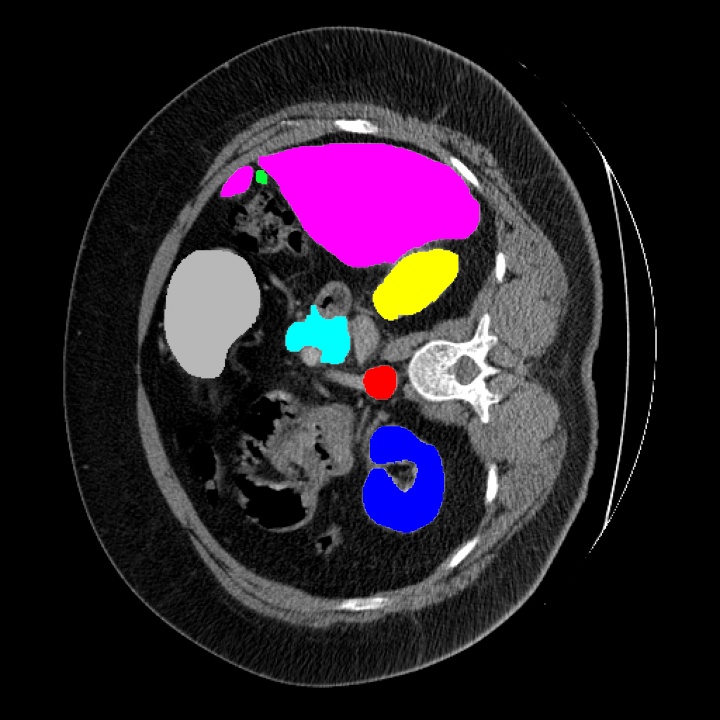}} &
    \vspace{0.15cm} \\   
    {\textbf{B}} &
    \raisebox{-.5\height}{\includegraphics[width=3cm]{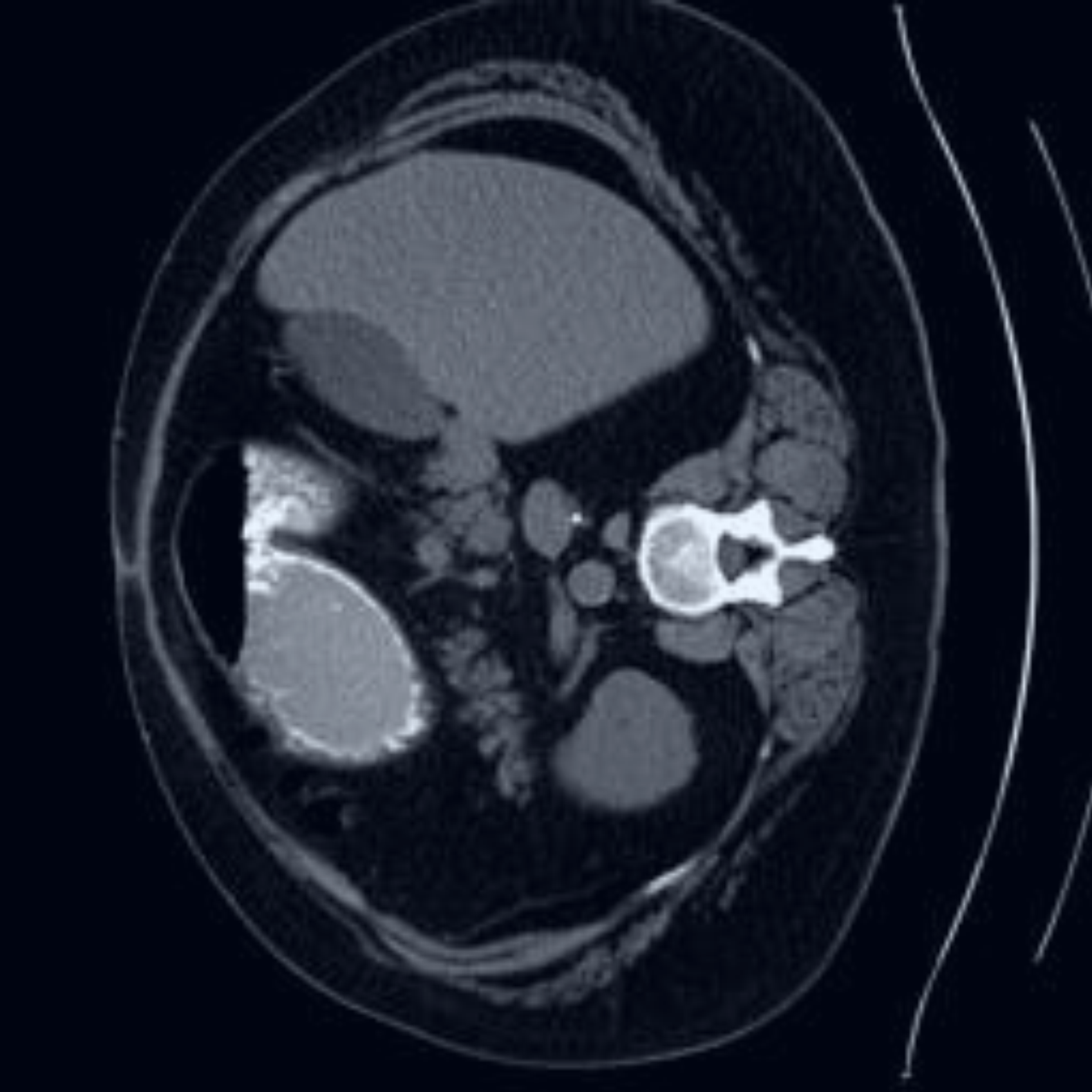}} &
 \raisebox{-.5\height}{\includegraphics[width=3cm]{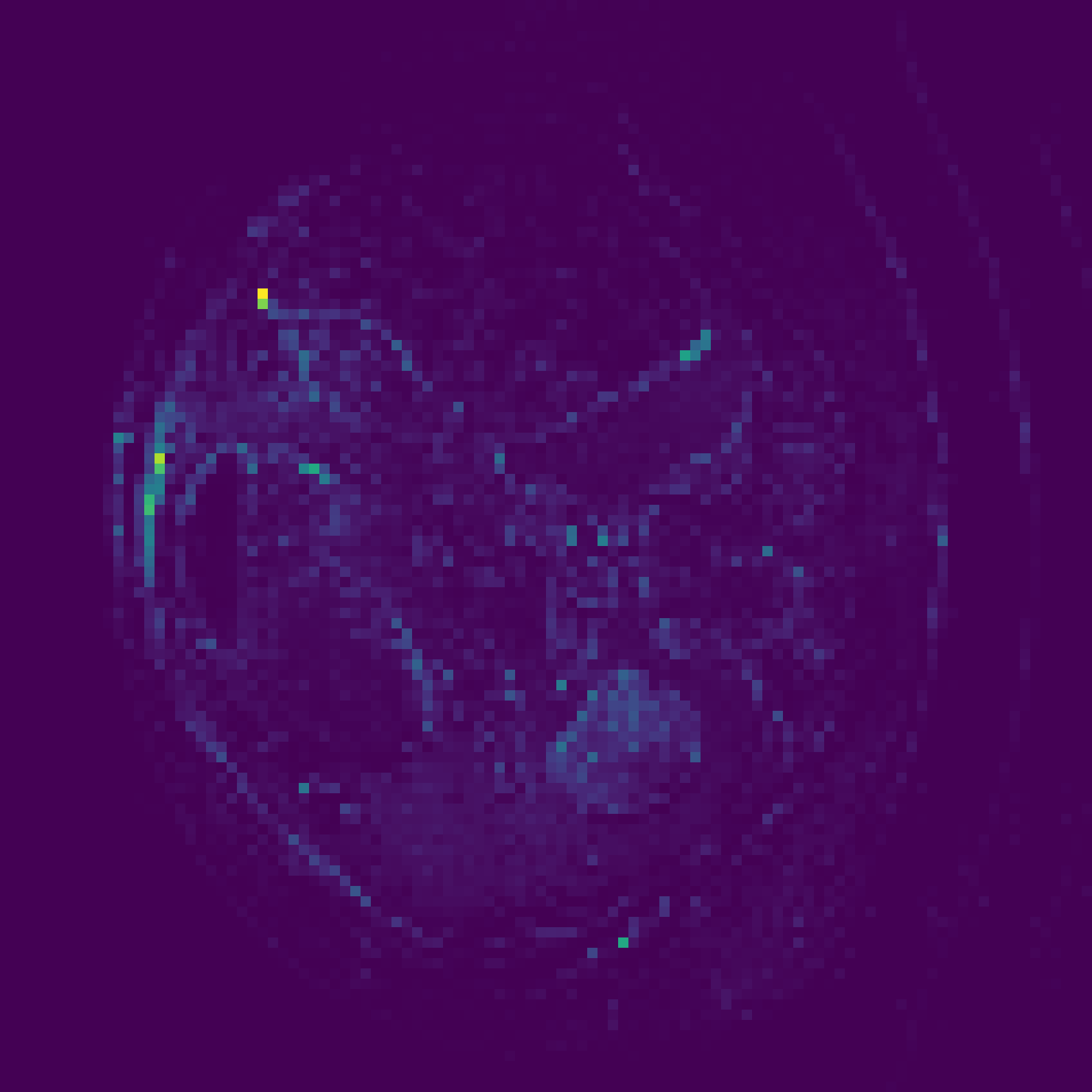}} &
  \raisebox{-.5\height}{\includegraphics[width=3cm]{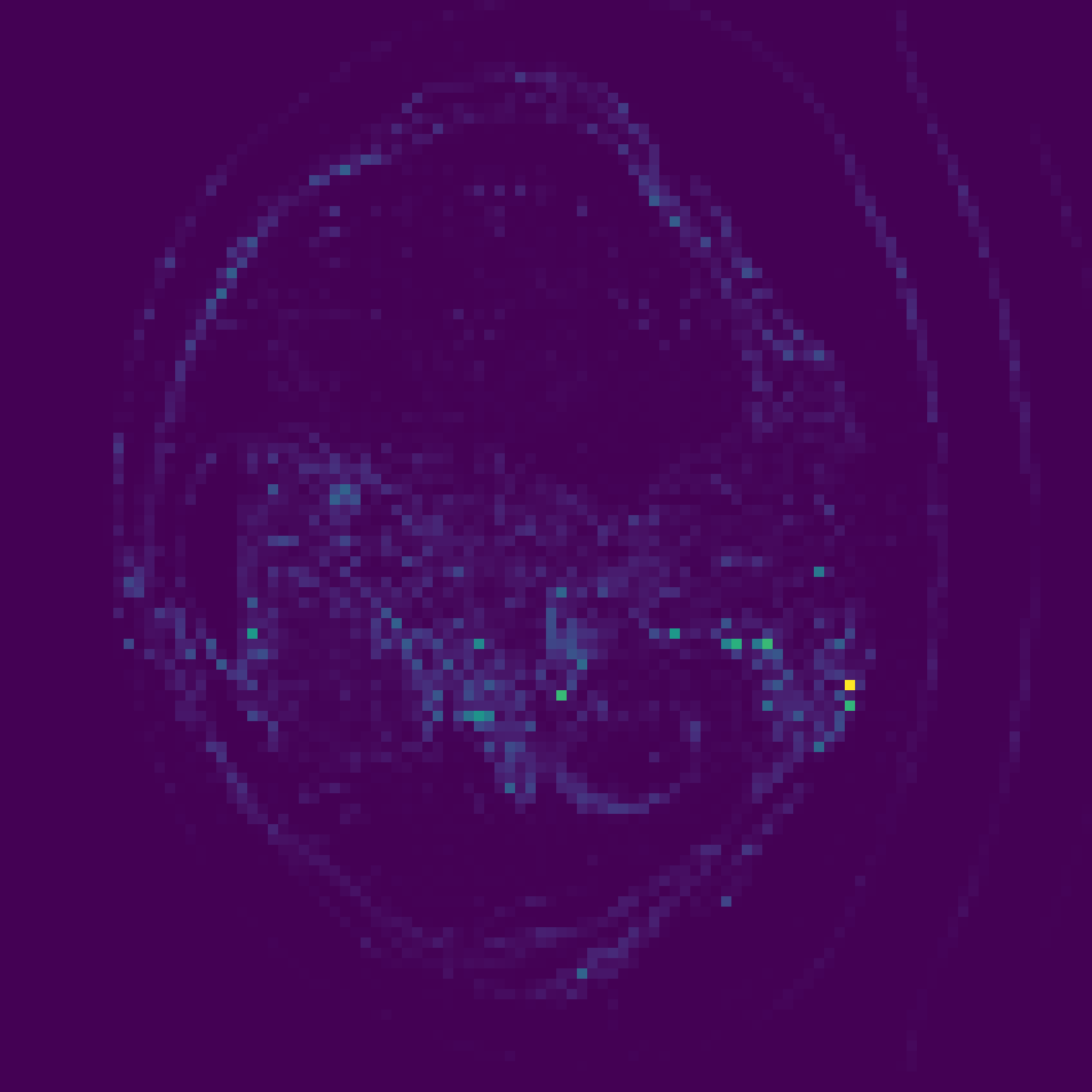}} &
   \raisebox{-.5\height}{\includegraphics[width=3cm]{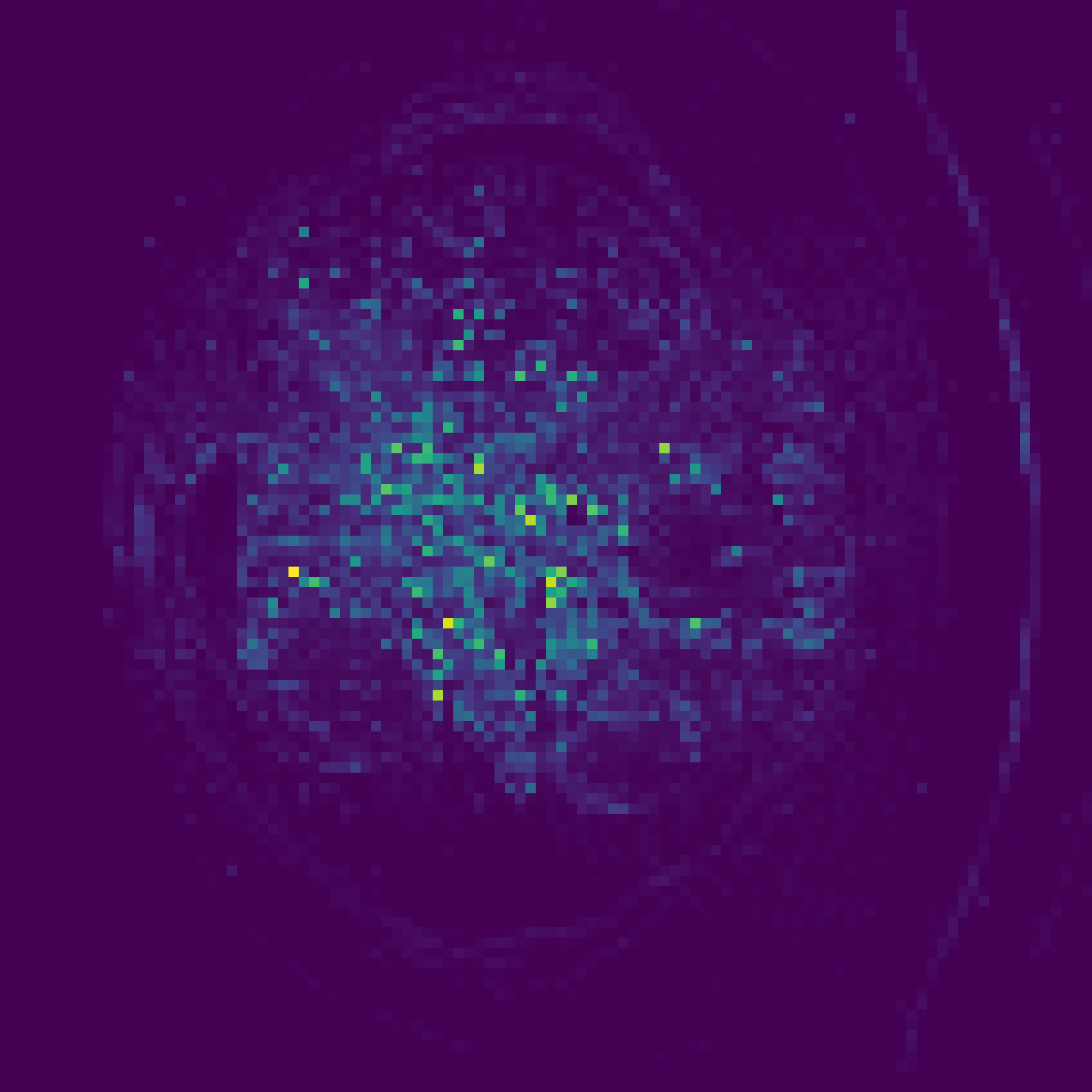}} &
    \raisebox{-.5\height}{\includegraphics[width=3cm]{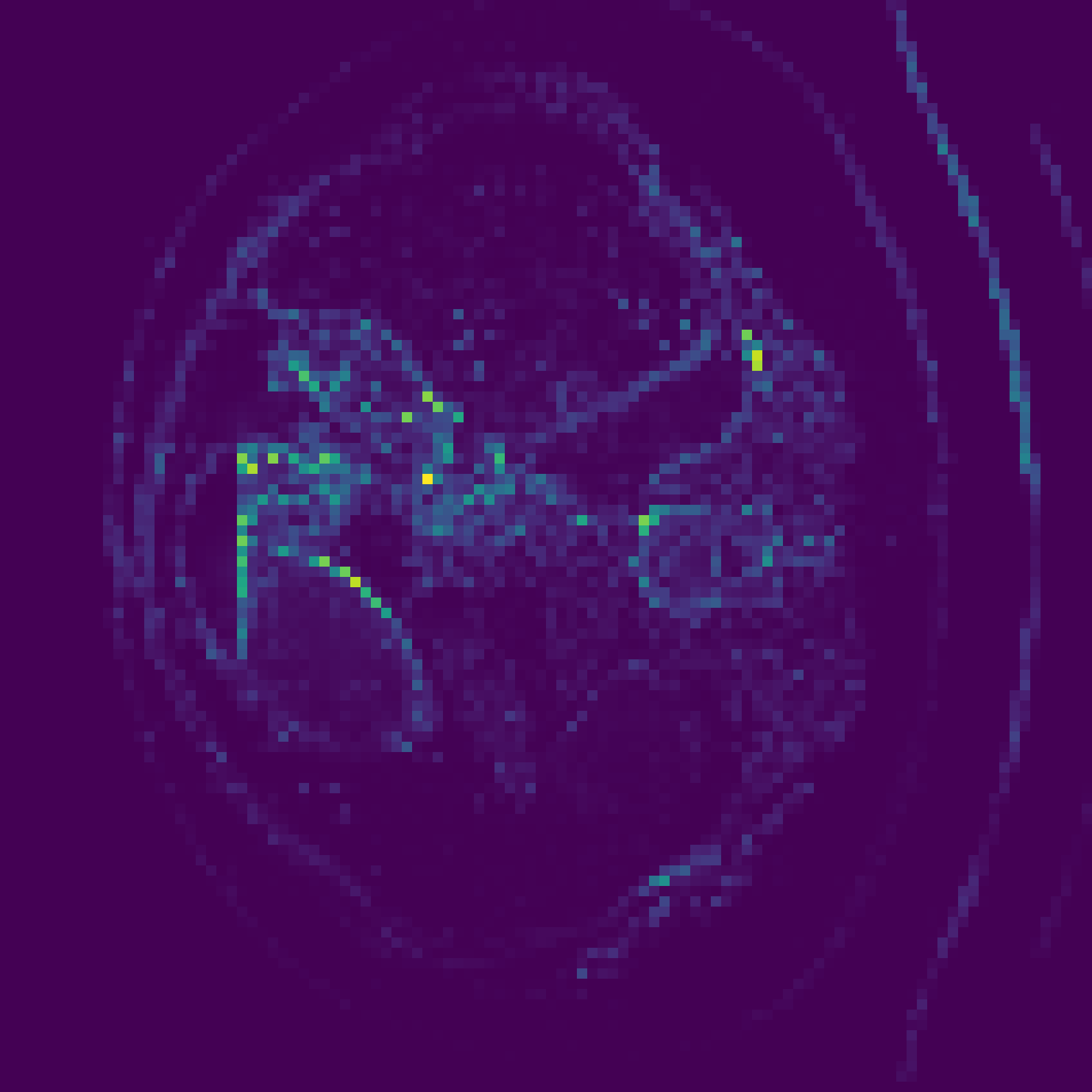}} &
     \raisebox{-.5\height}{\includegraphics[width=3cm]{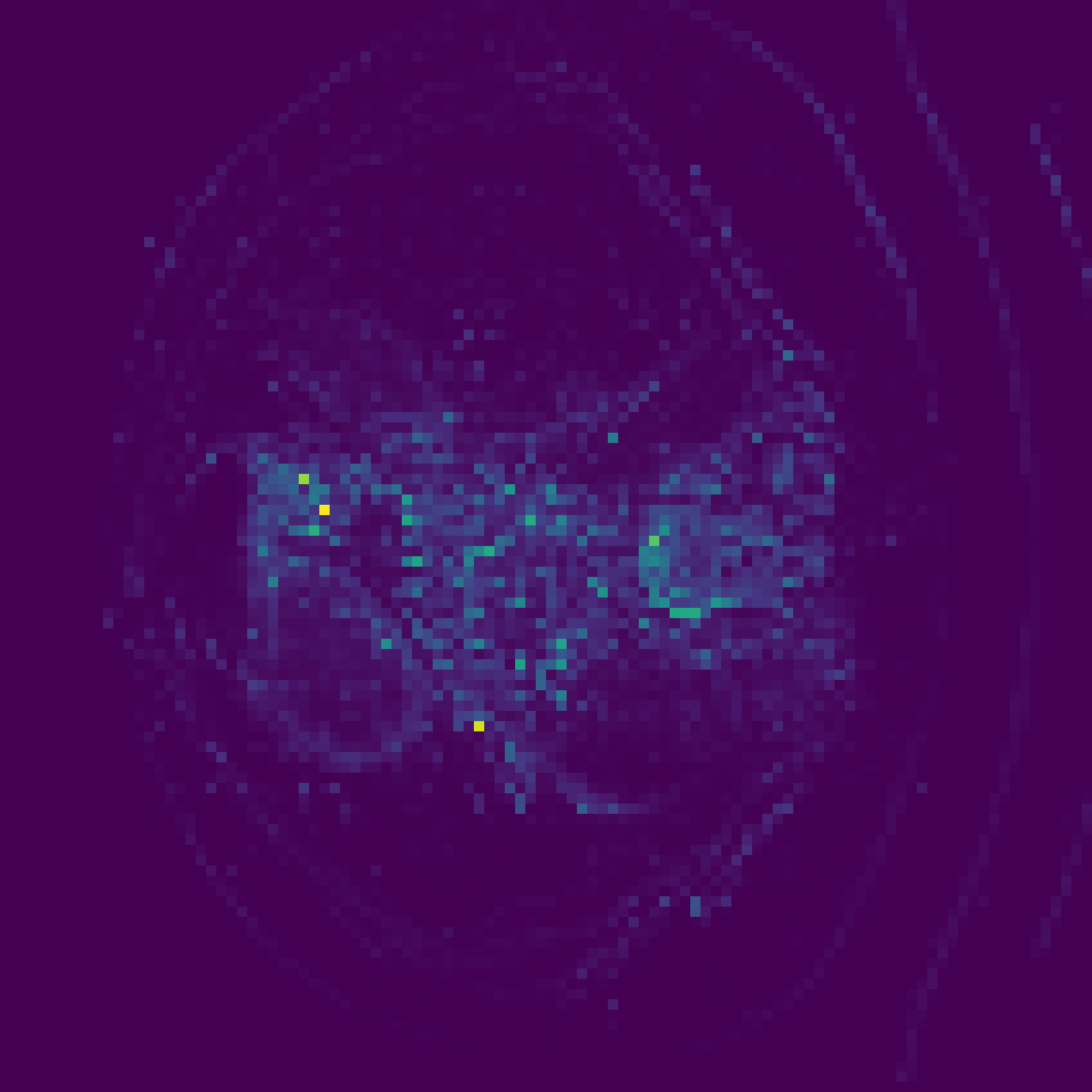}} &
      \raisebox{-.5\height}{\includegraphics[width=3cm]{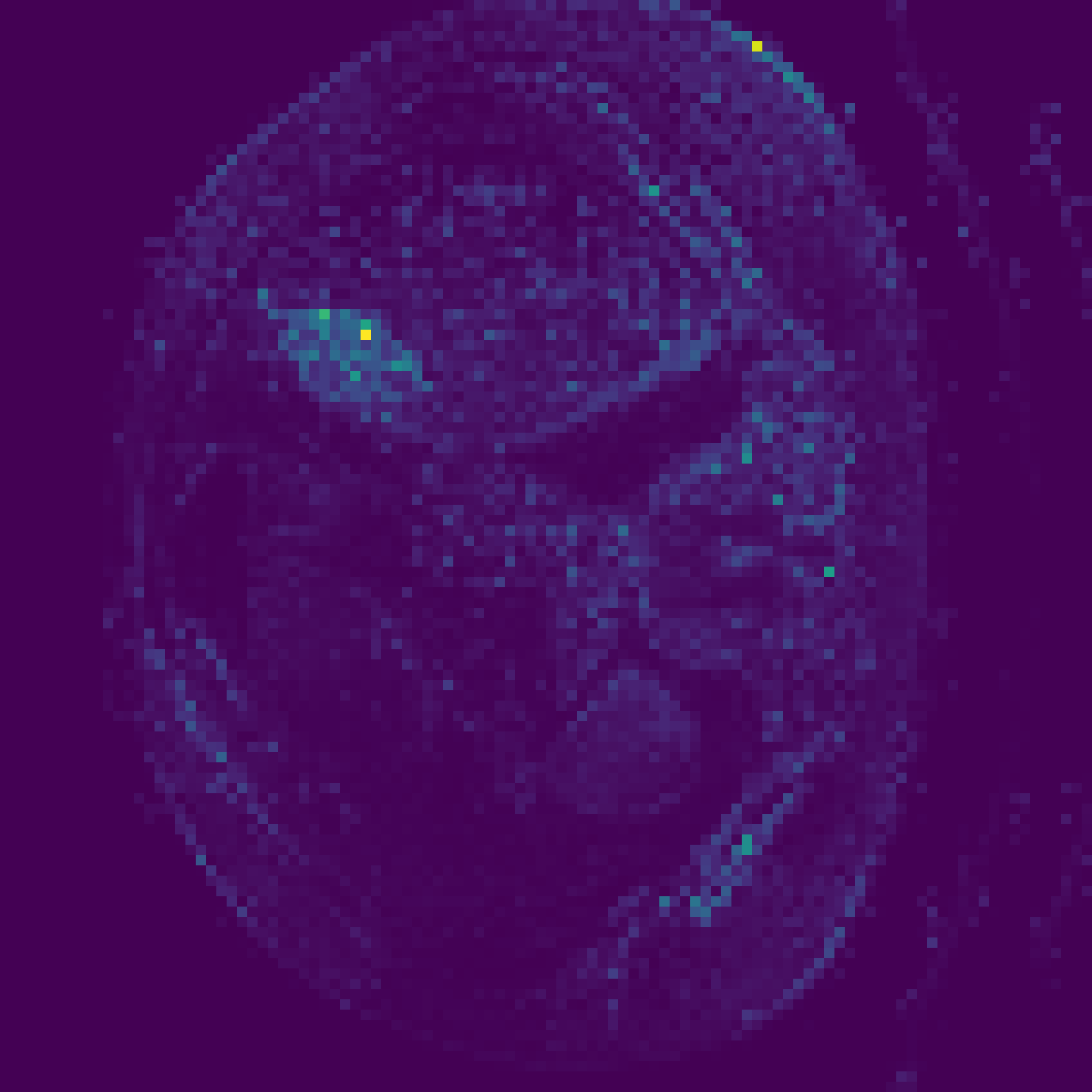}} &
 \raisebox{-.5\height}{\includegraphics[width=3cm]{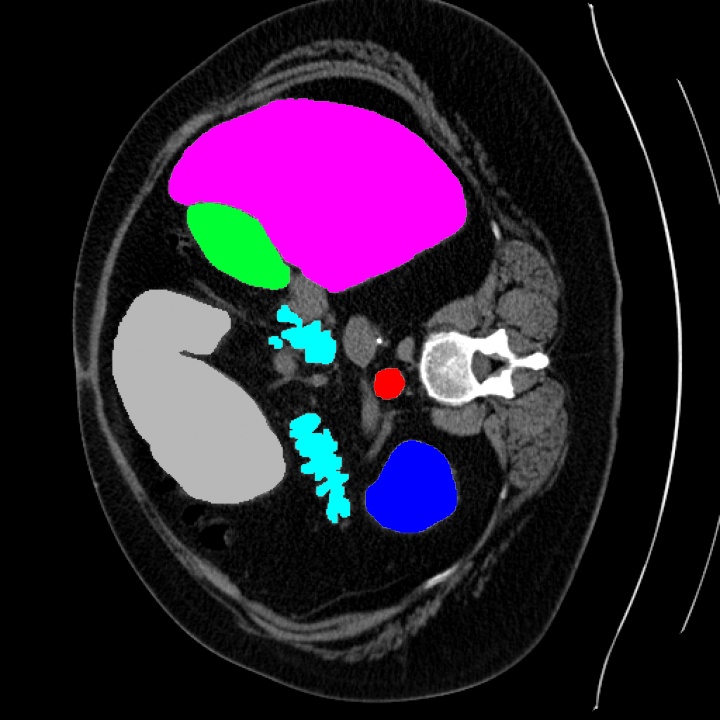}} &
    \vspace{0.15cm} \\  
 \end{tabular}
}

\captionof{figure}{\textbf{Attention Map Visualization.} Given an image, we generate the attention map visualizations using a pretrained DINO \cite{caron2021emerging}. The attention maps bear observable resemblance to the ground truth segmentation maps. Inpired by this phenomenon, we innovatively pass these segmentation map visualizations along with the input image to the segmentation network to generate the semantic segmentation prediction.  }
\label{ch:res:visualization}

\end{center}
\end{strip}

\begin{abstract}

Although purely transformer-based architectures showed promising performance in many computer vision tasks, many hybrid models consisting of CNN and transformer blocks are introduced to fit more specialized tasks. Nevertheless, despite the performance gain of both pure and hybrid transformer-based architectures compared to CNNs in medical imaging segmentation, their high training cost and complexity make it challenging to use them in real scenarios. In this work, we propose simple architectures based on purely convolutional layers, and show that by just taking advantage of the attention map visualizations obtained from a self-supervised pretrained vision transformer network (e.g., DINO \cite{caron2021emerging}) one can outperform complex transformer-based networks with much less computation costs. 
The proposed architecture is composed of two encoder branches with the original image as input in one branch and the attention map visualizations of the same image from multiple self-attention heads from a pre-trained DINO model (as multiple channels) in the other branch. 
The results of our experiments on two publicly available medical imaging datasets show that the proposed pipeline outperforms U-Net and the state-of-the-art medical image segmentation models.

\end{abstract}

\section{Introduction}
\label{sec:intro}
Medical image segmentation aims to highlight critical parts of an image, such as organs and tumors, in various modalities (e.g., CT, MR, etc), for the purpose of clinical diagnosis. The cost and availability of annotation and segmentation by experts necessitates automatic methods, and deep-learning based techniques have given a significant boost to the field in recent years. The next leap in performance is likely to stem from 
methods that can leverage large amounts of unannotated data in a self-supervised manner, and uncover the underlying structure and knowledge in the data to improve segmentation of medical images.

The pipelines in medical image segmentation networks commonly follow an encoder-decoder architecture resembling a pixel-to-pixel mapping from image to segmentation maps. 
One of the earliest and most successful representatives of this approach is U-Net \cite{ronneberger2015u}, which uses CNN blocks and skip-connections from encoder to decoder at different resolution levels. 
Many works proposed CNN architectures that were inspired by or expanded upon U-Net \cite{he2016deep, szegedy2015going, simonyan2014very, huang2017densely}. Until today though, most CNN-based approaches are outperformed by ensembles of vanilla U-Net architectures as proposed in nnU-Net \cite{isensee2020nnunet}, which showed that image pre-processing heuristics may have a larger effect than architectural improvements. The inductive bias of CNNs is beneficial for faster convergence. However, it can also cause the model to saturate faster and miss complex underlying relations. Therefore, more complex networks like attention-based models \cite{chen2021transattunet} were explored for medical image segmentation.

Transformers \cite{vaswani2017attention} introduced attention layers for natural language processing (NLP), and were recently adapted to Vision Transformers (ViT) for different computer vision tasks \cite{dosovitskiy2020imageVIT}. 
Some models for medical image segmentation employed purely transformer-based architectures \cite{chen2021transunet,cao2021swinunet}. Others followed a hybrid approach and incorporated transformer blocks in the encoder \cite{chang2021transclaw, yao2021claw,xu2021levit,chen2021transunet}, decoder \cite{li2021medical,li2021more}, both \cite{wang2020linformer,zhou2021nnformer,lin2021ds} or other network parts \cite{chen2021transattunet,yancy2021form,lambert2020segthor}. 
Although transformer-based designs lead to a lack of inductive bias and a wider field of view compared to CNNs, 
they require a large amount of data for training. Further, the memory requirement of attention layers grows quadratically with the number of image patches, which leads to higher computational resources compared to CNNs, particularly during training. Interpretability is also another crucial criteria, especially in sensitive scenarios like medical applications, and despite some efforts to improve it\cite{chefer2021transformer,kim2022vit}, investigation of interpratibility of attention-based architectures is more challenging compared to CNNs\cite{jain2019attention}.

The lack of carefully annotated data in medical imaging is another challenge to adopt transformers for medical image segmentation.{}
As a consequence, the adoption of self-supervised learning methods for segmentation tasks has been proposed \cite{cao2021swinunet,chen2021transunet}. There are several approaches for pretraining of Transformers. The most common one is introduced for NLP through defining pretext tasks \cite{vaswani2017attention}. Although adopting the same approach for images showed promising results \cite{dosovitskiy2020imageVIT},another approach recently adapted in self-supervised training for vision transformers \cite{caron2021emerging} could be a better fit for computer vision. It follows a student-teacher scheme to distill the knowledge of the teacher branch to the student branch. The teacher branch receives a larger field of view, and the knowledge of teacher distill the student in an unsupervised manner.Despite different approaches of training, both DINO \cite{caron2021emerging} and ViT \cite{dosovitskiy2020imageVIT} reported an emergence of meaningful shapes in their attention map visualizations. Interestingly, both DINO \cite{caron2021emerging} and ViT \cite{dosovitskiy2020imageVIT} reported an emergence of meaningful shapes in their attention map visualizations. In this work, we used the pretrained DINO  model for the generation of pairs of image-attention map visualization.

To leverage the benfits of transformers in medical image segmentation without having to bear the limitations and high complexities, we propose to generate pairs of image-attention map visualizations to combine the perspective and richness of extracted features in transformers with the simplicity and effectiveness of CNNs in medical image segmentation. It is observed in \autoref{ch:res:visualization} that each head resembles certain parts of the segmentation maps, so the capturing of all of the available information is enforced by incorporation of the generated attention-map visualizations as input data. Based on these generated maps, we present two novel architectures: (i) The simpler one investigates the effectiveness of combining image and the generated attention map visualizations as one input, i.e., the main image as the first channel and the attention map visualizations as other channels of the input data; (ii) The more complex one considers two separate branches for image and attention-map visualization, such that all visualization maps are considered as channels of one input image. We show that the latter architecture outperforms the first one as well as many other prominent methods of medical image segmentation.

In summary, our main contributions in this work are as follows:
\begin{itemize}
\item We demonstrate that transformers trained in a self-supervised manner could capture essential information, which to the best of our knowledge is the first work that directly incorporate them for image segmentation.
\item  We introduce simple yet effective CNN-based architectures to employ the attention maps visualizations along the original image for medical image segmentation, which potentially allows for more investigatory measures like interpretability that has been widely studied for CNNs and is cruical for medical applications. 
\item We show the effectiveness of our approach in two well-known datasets and compare its performance to other well-known segmentation techniques based on transformers and CNNs.
\end{itemize}

\section{Related Work}
\label{sec:related}
Deep learning has immensely affected medical image segmentation. Among earlier architectures, Long \etal \cite{long2015fully} introduced an image-image mapping based on Fully Convolutional Network. Later, U-Net revolutionized medical image segmentation with its encoder-decoder architecture connecting in a bottleneck and skip-connections between encoder and decoder components \cite{ronneberger2015u}; since then, U-Net has been primarily used as a benchmark in medical image segmentation. Some other architectures also proposed incorporating the components of U-Net, e.g., \cite{he2016deep, szegedy2015going, simonyan2014very, huang2017densely}, which most of them use U-Net-based skip connections. Generalization is improved by shortening the gap between the encoder and decoder semantic maps~\cite{zhou2018unet++}. Adding residual connections for each encoder and decoder block and dividing the input image into patches with a weighted map for each patch as input to the model is also investigated~\cite{xiao2018weighted}. Diakogiannis \etal \cite{diakogiannis2020resunet} utilized U-Net backbone, combined with residual connections, atrous convolutions, pyramid scene parsing pooling and multi-tasking inference, and Jha \etal \cite{jha2019resunet++} explored the advantages of residual blocks combined with the squeeze and excitation blocks.
V-Net \cite{milletari2016v} and 3D U-Net \cite{3dunet} utilized similar architecture for 3D medical image segmentation. U-Net architecture use-cases also expanded to other medical image analysis tasks such as computer-aided diagnosis \cite{tang2018automated,tang2019nodulenet,tang2019end,liao2019evaluate,ding2017accurate}, image denoising \cite{you2019low,lyu2018super}, image registration \cite{balakrishnan2019voxelmorph,heinrich2019closing}, and it is also utilized in diffusion models for state-of-the-art image generation \cite{ho2020denoising}. Apart from the traditional U-Net form, Y-Net \cite{farshad2022upsilon}, a segmentation network with two encoders and one shared decoder has been recently proposed for medical image segmentation. Y-Net introduces a spectral encoder that extracts frequency domain features and shows superior performance. The spectral encoder consists of Fast Fourier convolutional (FFC) blocks with fast Fourier transform at their core.

Convolutional layers have inherent inductive bias, that limit the ability of the network to make spatial relation in different parts of the input image. Some works employed sequence-to-sequence modules like RNN and LSTM  in medical image segmentation \cite{hesamian2019deep,shahzadi2018cnn}. Dilated convolutions \cite{yu2015multi}, as the name suggests, tried to modify convolutional layers to have a broader view, and they were used by many methods to preserve the spatial size of the feature map and to enlarge the receptive field~\cite{chen2014semantic,  zhao2017pyramid, zhang2018context}. The receptive field can also be enlarged by using larger kernels~\cite{peng2017large} and enriched with contextual information by using kernels of multiple scales~\cite{chen2017deeplab}. Hardnet~\cite{chao2019hardnet} employs a Harmonic Densely Connected Network that is shown to be highly effective in many real-time tasks, such as classification and object detection. Wang \etal used attention maps in each feature map of the encoder-decoder block to enlarge the representation between farther areas in the image \cite{wang2020axial} and it was extended by the incorporation of scalar gates in the attention layer \cite{valanarasu2021medical}. In the same work, they proposed Local-Global training strategy or LoGo that aims to consider specialized branches for local and global views.

\subsection{Vision Transformers}

Transformer architecture was first introduced for machine translation and thereafter became the dominant backbone architecture for Natural Language Processing; it is constructed mainly by attention-blocks in the form of encoder-decoder, which utilized the GPUs much more efficiently compared to other sequence-sequence blocks like LSTM and RNN \cite{vaswani2017attention}. Depending on the task, many architectures adopted the original encoder-decoder form \cite{lewis2019bart,lample2019cross,conneau2019unsupervised} and some only used the encoder part \cite{devlin-etal-2019-bert,liu2019roberta,kitaev2020reformer,beltagy2020longformer} or decoder part \cite{radford2018improving,dai2019transformer,yang2019xlnet}. Inspired by the success of transformers, many attempted to bring transformers to vision tasks \cite{Wang2018,cao2019gcnet,ramachandran2019stand,zhao2020exploring}, and Dosovitskiy \etal \cite{dosovitskiy2020imageVIT} inspired by the relative simplicity of BERT architecture \cite{devlin-etal-2019-bert}, introduced vision transformers (ViT) as the state-of-the-art in image classification tasks; They also explored the concept of hybrid architectures that uses CNNs to generate embeddings, and other works like CvT \cite{wu2021cvt} improved the ViT performance by incorporating CNNs. LeViT \cite{graham2021levit} used low-resolution attention maps and combined them with CNNs to improve the inference time. Many other works extend ViT to other tasks like video classification \cite{timesformer} and ViLT \cite{kim2021vilt} extract and combine features from text and image for better inference. To fit the transformer architecture for tasks other than classification, hybrid architectures were proposed. For object detection, DETR \cite{detr} uses a CNN block as the entry backbone for feature extraction and bounding box detection and uses the encoder-decoder transformer block for prediction of the labels. For semantic segmentation, SETR \cite{SETR} transformers are implemented to extract features in the encoder for sequentializing images without using the traditional FCN \cite{long2015fully}. Ranftl \etal \cite{ranftl2021vision} leverage vision transformers in place of CNNs as a backbone for dense prediction tasks.
\subsubsection{Self-Supervised Learning in Vision Transformers}
Self-supervised learning to pretrain transformers initially introduced as pretext task in NLP by masking random embedding vectors and optimizing the model to recover them. In the same way, ViT \cite{dosovitskiy2020imageVIT} also uses a simple pretext approach, by randomly masking patches of the input image, and asking the model to predict its average color. In another work, Atito \etal \cite{atito2021sit} utilize group mask model learning (GMML) for pretext pretraining of vision transformers to improve the simple masking approach. SelfPatch \cite{yun2022patch} is also a pretext technique that aims to learn better patch-level representations. On the other hand, DINO \cite{caron2021emerging} follows a student-teach scheme that is more common in computer vision. EsViT \cite{li2021efficient} also exploits Knowledge Distillation with a fixed teacher network and a student network that is continuously updated in an attempt to minimize a loss function.
Caron \etal \cite{caron2021emerging} claimed in DINO that with this self-supervised approach, the extracted features in self-supervised vision transformers contain meaningful and intuitive information about the image, which does not appear as explicit as supervised vision transformers or CNNs, which also inspired this work to benefit from such information in medical image segmentation. Ge \etal \cite{ge2021revitalizing} also presented a pipeline to use the priors of transformers in a CNN network; it has two branch architectures: one for the CNN, and the other for the transformer that guides the CNN branch in a self-supervised manner, and both branches are trained simultaneously.

\subsection{Transformers for medical image segmentation}
A pure transformer-based model for medical image segmentation was introduced by Karimi \etal \cite{chen2021transunet} for 3D medical image segmentation, and Swin-UNet \cite{cao2021swinunet} based on Swin-Transformers \cite{liu2021swin} showed its effectiveness on ACDC \cite{bernard2018deepACDC} and other datasets; however, hybrid architectures were more researched. In most of such architectures, similar to U-Net, an encoder-decoder shape is followed. Transformer blocks are utilized in different parts of such networks. Swin UNETR \cite{tang2021self}, Trans Claw-UNet \cite{chang2021transclaw}, Claw-UNet \cite{yao2021claw} and LeViT-UNet \cite{xu2021levit} adopted transformers for feature extraction in the encoder. TransUNet \cite{chen2021transunet} with a similar approach showed superior performance on synapse dataset \cite{synapsedataset} and ACDC \cite{bernard2018deepACDC}.  Fewer works explored Transformers in only decoder parts \cite{li2021medical,li2021more}. On the other hand, UTNet \cite{wang2020linformer}, nnFormer \cite{zhou2021nnformer} and Dual Swin Transformer UNet (DS-TransUNet) \cite{lin2021ds} incorporated the transformers along CNNs in both encoder and decoder part. TransAttUNet \cite{chen2021transattunet} employed guided attentions in skip-connections to provide more expressive representation. Axial Fusion Transformer UNet (AFTer-UNet) \cite{yancy2021form} implement a fusion layer with axial fusion layers, and SegTHOR \cite{lambert2020segthor} also suggested another type of fusion layer.

\section{Method}
\label{sec:methods}
\begin{figure*}[!t]
\centering
\includegraphics[width=\linewidth]{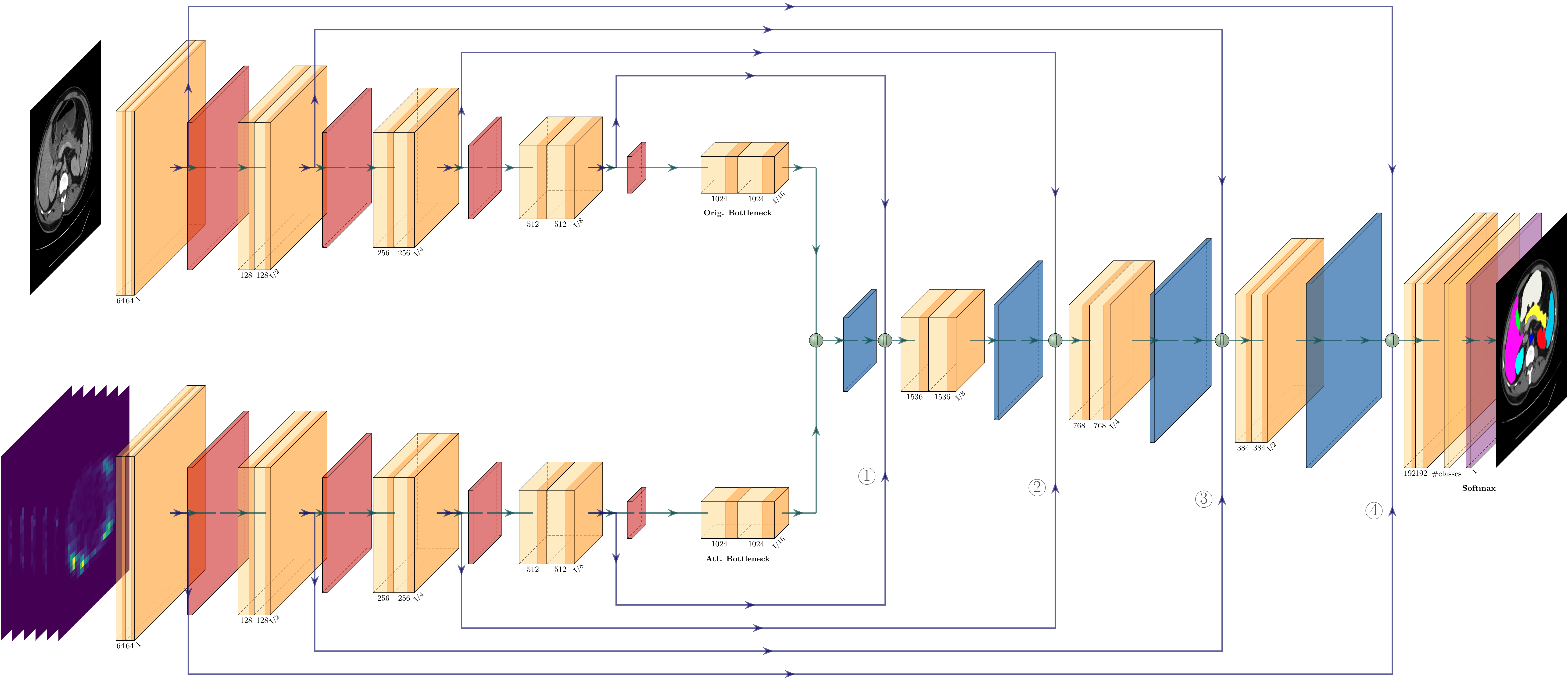}
\caption{\textbf{\methodName{} Architecture.} (\begingroup\setlength{\fboxsep}{0pt}\fcolorbox{black}{ConvColorLegend}{\rule{0pt}{7pt}\rule{7pt}{0pt}}\endgroup~Convolution, \begingroup\setlength{\fboxsep}{0pt}\fcolorbox{black}{ConvReluColorLegend}{\rule{0pt}{7pt}\rule{7pt}{0pt}}\endgroup~ReLu + Batch Norm, \begingroup\setlength{\fboxsep}{0pt}\fcolorbox{black}{PoolColorLegend}{\rule{0pt}{7pt}\rule{7pt}{0pt}}\endgroup~Max. Pooling, \begingroup\setlength{\fboxsep}{0pt}\fcolorbox{black}{SoftmaxColorLegend}{\rule{0pt}{7pt}\rule{7pt}{0pt}}\endgroup~Softmax, \begingroup\setlength{\fboxsep}{0pt}\fcolorbox{black}{UnpoolColorLegend}{\rule{0pt}{7pt}\rule{7pt}{0pt}}\endgroup~Transposed Convolution, \begingroup\setlength{\fboxsep}{0pt}\fcolorbox{black}{ConcatColorLegend}{\rule{0pt}{7pt}\rule{7pt}{0pt}}\endgroup~Concatenation.)} 
\label{ch:methods:ynetfig}
\end{figure*}
We present a simple architecture for semantic segmentation adapted from the well-known U-Net \cite{ronneberger2015u}. Unlike other complicated methods, rather than incorporating transformer components in our architecture, we propose extracting attention map visualizations from a pre-trained vision transformer model and feeding them to the segmentation network in addition to the input image. Our framework consists of two steps: 1) self-supervised training of the DINO model to obtain the attention map visualizations for images, 2) Training our proposed segmentation model using the segmentation map annotations, input images, and their corresponding attention map visualizations from step 1. \autoref{ch:methods:ynetfig} shows an overview of our proposed segmentation network in step (2), while the input to the lower branch is obtained from step (1).

\subsection{Definitions}
Given a dataset $\mathcal{D}$ of input images $x \in \mathbb{R}^{H \times W \times C}$ with height $H$, width $W$ and number of input channels $C$, and their corresponding segmentation map annotations $y \in \mathbb{R}^{H \times W \times N}$, where $\{x, y\} \in \mathcal{D}$. The number of classes in our dataset is defined by $N$. The goal of our segmentation model parameterized by $\theta$ is to predict the semantic segmentation maps $\hat{y} = \theta(x)$. We denote the pre-trained DINO \cite{caron2021emerging} model by $\phi$. We extract the attention map visualization from each self-attention head in DINO and denote them by $\nu_i$ where $i \in \{1,...,h\}$ defines the head index and $h$ defines the total number of heads. Therefore, the predicted segmentation map becomes a function of $\nu$ in addition to $x$, leading to $\hat{y} = \theta(x, \nu)$.

\subsection{DINO}
As mentioned earlier, DINO \cite{caron2021emerging} is a vision transformer-based architecture that is trained in a self-supervised manner without labels. DINO consists of a teacher (parameterized by $\phi_t$) model and a student model (parameterized by $\phi_s$) and is trained using self-distillation. The teacher and the student share the same architecture and receive two augmented cropped views $x_1$, $x_2$ from the input image $x$. The architecture of $\phi$ is based on a vision transformer \cite{dosovitskiy2020imageVIT} (ViT), followed by a projection head with the dimension $K$ that outputs the probability distributions $P_s , P_t$ of the student and teacher models, respectively; where $P ( x ) ^ { ( j ) }$ is the probability distribution of input image patch $x ^ {(j)} $ in \autoref{eq:dinoprob}:
\begin{equation} \label{eq:dinoprob}
P ( x ) ^ { ( j ) } = \frac { \operatorname { exp } ( \phi( x ) ^ { ( j ) } / \tau) } { \sum _ { k = 1 } ^ { K } \operatorname { exp } ( \phi ( x) ^ { ( k ) } / \tau ) }.
\end{equation}

Each network has its own parameters $\phi_s, \phi_t$ and temperature $\tau_s, \tau_t$, which yields $P_s, P_t$ using \autoref{eq:dinoprob}. The temperature parameter defines the sharpness of the probability distribution $P$. 

The backbone model (ViT-S/16) receives a grid of image patches $x ^ {(j)}$ with resolution $16 \times 16$ as input. A set of embeddings are generated by feeding the image patches to a linear layer, and are then followed by a learnable token with the goal of aggregating the information from the whole grid sequence. The embeddings are passed to a standard Transformer network which is a sequence of self-attention and feed-forward layers with skip connections. There are in total $h$ self-attention heads in the network, that generate the attention map visualizations ${\nu_{i,j}}$ for $i \in ({1,...,h})$ of image patch $x ^ {(j)}$.

\paragraph{DINO Optimization}
The student model parameters are updated by applying stochastic gradient descent and minimizing the cross-entropy loss between the features from the student and teacher (\autoref{eq:stu}):
\begin{equation} \label{eq:stu}
\mathcal{L}_{st} = \frac{- P _ { t } ( x_1 ) \operatorname { log } P _ { s } ( x_2 )}{2} + \frac{- P _ { t } ( x_2 ) \operatorname { log } P _ { s } ( x_1 )}{2}.
\end{equation}

The teacher model parameters are optimized via an exponential moving average of the students' parameters using: 

\begin{equation}\label{eq:teacher}
    \phi_t \gets \lambda \phi_t + (1-\lambda \phi_s),
\end{equation}
where $\lambda$ defines a a cosine scheduler from $0.996$ to $1$. 

\subsection{Semantic Segmentation}
The main contribution of our method lies in showing that the self-attention visualizations in a self-supervised pretrained transformer-based architecture can be employed without actually using the transformer components in the main model architecture, thus leading to a simple architecture, similar to U-Net, that uses additional features as input to perform the segmentation task more effectively. We also define a switching mechanism that allows further customization for keeping or removing skip connections.

We present two versions of our segmentation model. The first version simply adapts the U-Net \cite{ronneberger2015u} model by increasing the number of input channels in $\theta$ and concatenating the input image $x$ with the attention map visualizations $\nu$. This modification is done by changing the number of input channels in the initial layer of the U-Net from $C$ to $C+h$ (number of attention heads).

We hypothesize that since the input image and the attention map visualizations are from different domains, introducing an extra encoder to extract the features from the attention map visualizations would lead to better feature modulation. Furthermore, we believe that the attention map visualizations from a model encode more valuable information than the original image, which would have higher correlation with the final segmentation map; thus facilitating the possibility of the model to assign higher weights to such features in a separate encoder.

Therefore, we propose a network with two encoders and a shared decoder for predicting the semantic segmentation map. The first encoder $E_x$ gets the original image $x$ as input, while the second encoder $E_\nu$, which has $h$ input channels, receives the attention map visualizations $\nu$ of the different heads. The features extracted by these two encoders are concatenated at the bottleneck and then fed to the shared decoder $D$.

Since the skip connections from the encoders provide a direct connection to the decoder features and, as a result, the predicted segmentation map, we explore the settings of employing the skip connections at different points. These are denoted by switches (\circled{1}, \circled{2}, \circled{3}, \circled{4}) in our experiments defining whether the skip connection exists at the specified point or not. \autoref{ch:methods:ynetfig} depicts the \methodName{} network with all switches $\circled{1}=\circled{2}=\circled{3}=\circled{4}=1$.

\paragraph{Losses}
To optimize our segmentation model, we employ the combined cross-entropy, and dice loss \cite{jadon2020survey}. The dice-coefficient loss (\autoref{eq:exp:dice}) has high flexibility towards class imbalance, while the cross-entropy loss (\autoref{eq:exp:ce}) helps with the curve smoothing \cite{jadon2020survey}. 

\begin{equation}\label{eq:exp:ce} 
\ell_{CE} (\hat{y}, y) = - \frac{1}{N} \sum_{n=0}^N y_{n} \log (\hat{y}_{n}),
\end{equation}

\begin{equation}\label{eq:exp:dice}
\ell_{DICE}(\hat{y}, y) = 1- \frac{2y\hat{y}+\epsilon}{y+\hat{y}+\epsilon},
\end{equation}
where $\epsilon$ is added to the numerator and denominator for numerical stability. The total loss (\autoref{eq:exp:loss}) for the segmentation model then is the sum of dice and cross-entropy loss:
\begin{equation}\label{eq:exp:loss}
\mathcal{L}_{seg} = \frac{1}{2} \ell_{CE} + \frac{1}{2}\ell_{DICE}.
\end{equation}
\section{Experiments}
\label{sec:experiments}
\begin{table*}[ht]
\caption[Results of our segmentation architectures compared to related work.]{Quantitative results of our segmentation model compared to SOTA on Synapse Dataset \cite{synapsedataset} (UNet~$\dagger$: Same configuration as Swin-Unet). Note that results are obtained from the original paper if no marking or exponent. We extracted results marked with $^*$ from \cite{chen2021transunet} and $^\#$ from \cite{cao2021swinunet}.}
\label{ch:res:synapse:finaltable}
\resizebox{\linewidth}{!}{%
\begin{tabular}{@{}l|cccccccccc@{}}
\toprule
\multicolumn{1}{c|}{\textbf{Method}} & \textbf{DSC (\%)} $\uparrow$& \textbf{HD95 (mm)} $\downarrow$ & \textbf{Aorta} & \textbf{Gallbladder} & \textbf{Kidney (L)} & \textbf{Kidney (R)} & \textbf{Liver} & \textbf{Pancreas} & \textbf{Spleen} & \textbf{Stomach} \\ \midrule
V-Net$^*$ \cite{milletari2016v}& 68.81 & - & 75.34 & 51.87 & 77.10 & 80.75 & 87.84 & 40.05 & 80.56 & 56.98 \\
DARR$^*$ \cite{fu2020domain} & 69.77 & - & 74.74 & 53.77 & 72.31 & 73.24 & 94.08 & 54.18 & 89.90 & 45.96 \\
R50 U-Net$^*$ \cite{alam2020automatic} & 74.68 & 36.87 & 84.18 & 62.84 & 79.19 & 71.29 & 93.35 & 48.23 & 84.41 & 73.92 \\
R50 AttnUNet$^*$ \cite{schlemper2019attention}& 75.57 & 36.97 & 55.92 & 63.91 & 79.20 & 72.71 & 93.56 & 49.37 & 87.19 & 74.95 \\
U-Net$^\#$ \cite{ronneberger2015u} & 76.85 & 39.70 & 89.07 & 69.72 & 77.77 & 68.60 & 93.43 & 53.98 & 86.67 & 75.58 \\
AttnUNet$^\#$ \cite{schlemper2019attention} & 77.77 & 36.02 & 89.55 & 68.88 & 77.98 & 71.11 & 93.57 & 58.04 & 87.30 & 75.75 \\ \midrule
R50 ViT$^\#$ \cite{dosovitskiy2020imageVIT}& 71.29 & 32.87 & 73.73 & 55.13 & 75.80 & 72.20 & 91.51 & 45.99 & 81.99 & 73.95 \\
ViT$^*$ \cite{dosovitskiy2020imageVIT} & 61.50 & 39.61 & 44.38 & 39.59 & 67.46 & 62.94 & 89.21 & 43.14 & 75.45 & 69.78 \\
TransUNet \cite{chen2021transunet}& 77.48 & 31.69 & 87.23 & 63.13 & 81.87 & 77.02 & 94.08 & 55.86 & 85.08 & 75.62 \\
TransClaw U-Net \cite{chang2021transclaw} & 78.09 & 26.38 & 85.87 & 61.38 & 84.83 & 79.36 & 94.28 & 57.65 & 87.74 & 73.55 \\
MT-UNet \cite{wang2021mixed} & 78.59 & 26.59 & 87.92 & 64.99 & 81.47 & 77.29 & 93.06 & 59.46 & 87.75 & 76.81 \\ \midrule
U-Net $\dagger$ & 79.52 & 33.99 & 89.64 & 69.73 & 82.79 & 77.26 & 93.50 & 61.71 & 84.15 & 77.36 \\
Swin-UNet \cite{cao2021swinunet} &79.13 & 21.55 & 85.47 & 66.53 & 83.28 & 79.61 & 94.29 & 56.58 & 90.66 & 76.60 \\
TransCASCADE \cite{rahman2023medical} & 82.68 & 17.34 & 86.63 & 68.48 & 87.66 & \textbf{84.56} & 94.43 & 65.33 & 90.79 & \textbf{83.52} \\ \midrule
\methodName{} (Ours) &	\textbf{84.26}	&	\textbf{13.79}	&	\textbf{89.66}	&	\textbf{72.47}	&	\textbf{87.89}	&	83.90	&	\textbf{95.34}	&	\textbf{67.61}	&	\textbf{93.74}	&	83.48 \\ \midrule 
\end{tabular}
}

\end{table*}
In this section we present the results of our experiments on two publicly available datasets for medical image segmentation namely the ACDC \cite{bernard2018deepACDC} and Synapse \cite{synapsedataset} datasets. We use two different well-known modalities: CT and MR medical imaging; these datasets also holds different sets of organs with different sizes and intensities with challenging structures and shapes to evaluate our proposed networks. As follows, we shortly refer to the public benchmarks used in this work, then we present the experimental setup and implementation details. Finally, we demonstrate the results of our experiments compared against state-of-the-art (SOTA) methods and after that an ablation study of different settings of our proposed method. We used the average dice score for the evaluation, which is the standard metric in medical image segmentation.

\subsection{Datasets}
\paragraph{ACDC}
The Automated Cardiac Diagnosis Challenge (ACDC) dataset \cite{bernard2018deepACDC} consists of cardiac cine MR images from 150 patients. We follow the same experimental protocol as MT-UNet \cite{wang2021mixed} and TransUNet \cite{chen2021transunet}, which utilize the data of 100 out of the 150 patients in this dataset. For each data sample in the dataset, the data for two modalities of end-diastole (ED) and end-systole (ES) are provided. The annotations provided in the dataset provide semantic segmentation maps belonging to three regions, left ventricle (LV), right ventricle (RV), and myocardium (Myo). The train / validation and test splits follow the same setting as previous work with 70, 10, and 20 samples respectively. The slices in this dataset have a resolution of $352\times352$ and each volume has between 7 and 17 slices. 

\paragraph{Synapse}
The Synapse dataset \cite{synapsedataset} is a multi-organ segmentation dataset of abdominal CT images. Similar to ACDC, we follow the same experimental protocol to MT-UNet \cite{wang2021mixed} for training and evaluation of our model. In total, 30 abdominal CT scans and their corresponding semantic segmentation maps, belonging to eight abdominal organs (aorta, gallbladder, spleen, left kidney, right kidney, liver, pancreas, spleen, and stomach) are adopted. Each CT volume in the dataset has varying number of between 85 to 198 slices with a resolution of $512\times512$. Similar to \cite{wang2021mixed}, we employ 18 volumes for training and 12 volumes for testing.

\subsection{Experimental Setup}
We follow the same training and evaluation scheme as \cite{wang2021mixed}, if not otherwise stated. We use the Adam optimizer \cite{DBLP:journals/corr/KingmaB14} for training the model. The initial learning rate is set to $1\mathrm{e}{-4}$, with a step scheduler to gradually decrease the learning rate in each iteration of the training with a rate of $0.9$. The model has a weight decay of $1\mathrm{e}{-4}$ for regularization. The maximum number of training epochs is set to $300$, while we apply early stopping for our model based on the validation dice score. The batch size was set to 16 and the image resolution was $224 \times 224$. For augmentation purposes, we apply random flipping and rotations to the images during the training. We report the dice score value for different organs and regions and the average dice score for both datasets. For the Synapse dataset \cite{synapsedataset}, we also report the Hausdorff distance (HD95) between the predicted and ground truth segmentation maps.

For the extraction of attention map visualizations for both ACDC and Synapse datasets, we employed the pre-trained DINO \cite{caron2021emerging} model on ImageNet and fine-tuned it on the corresponding dataset. We adopt the ViT-S/16 model and fine-tune it with a batch size of 64 for 800 epochs and with a learning rate of $1\mathrm{e}{-4}$ on the training set of each dataset. The DINO model is fine-tuned on images with a resolution of $256 \times 256$. To employ attention map visualizations with the same resolution as the input images for the segmentation task, we downsample the attention map visualization to $224 \times 224$. The ViT-S/16 model has $6$ attention heads, which would also set the number of input channels to our attention encoder to $6$.

\subsection{Results}
We present quantitative and qualitative results of our method compared against the state-of-the-art as follows. Then, we show an ablation study of the components of our architecture. The quantitative results of our model compared to the SOTA on the ACDC dataset is presented in \autoref{ch:res:table:acdc}. All the experiments demonstrate the superiority of \methodName{} to comparable previous work. \methodName{} outperforms the complex transformer-based architectures such as MT-UNet \cite{wang2021mixed} and Swin-UNet \cite{cao2021swinunet} by a large margin in terms of dice score. The qualitative results in \autoref{ch:res:fig:synapse:figures:segperf} show that \methodName{} predicts the segmentation map more accurately compared to U-Net. Also, \methodName{} is able to segment small structures and wholes more effectively.
\subsubsection{Comparison to SOTA}
\begin{figure}[ht]
\centering
\resizebox{\linewidth}{!}{
 \begin{tabular}{c@{\hskip 0.15cm}c@{\hskip 0.15cm}c@{\hskip 0.15cm}}
\textbf{Ground Truth} & \textbf{U-Net} \cite{ronneberger2015u} & \textbf{\methodName{} (Ours)} \\
 \raisebox{-.5\height}{\includegraphics[width=3cm]{figures/comparison/unet/case0025_gt_54.jpg}} &
 \raisebox{-.5\height}{\includegraphics[width=3cm]{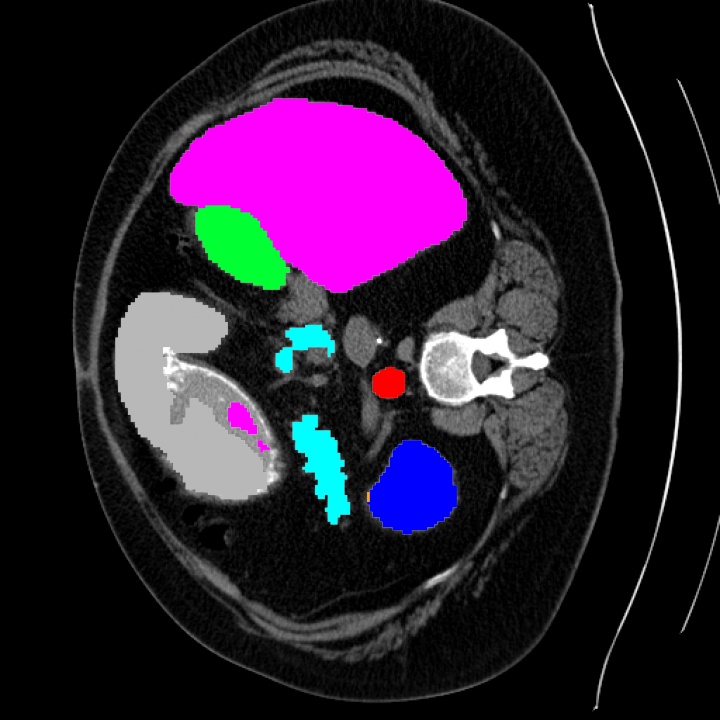}} &
    \raisebox{-.5\height}{\includegraphics[width=3cm]{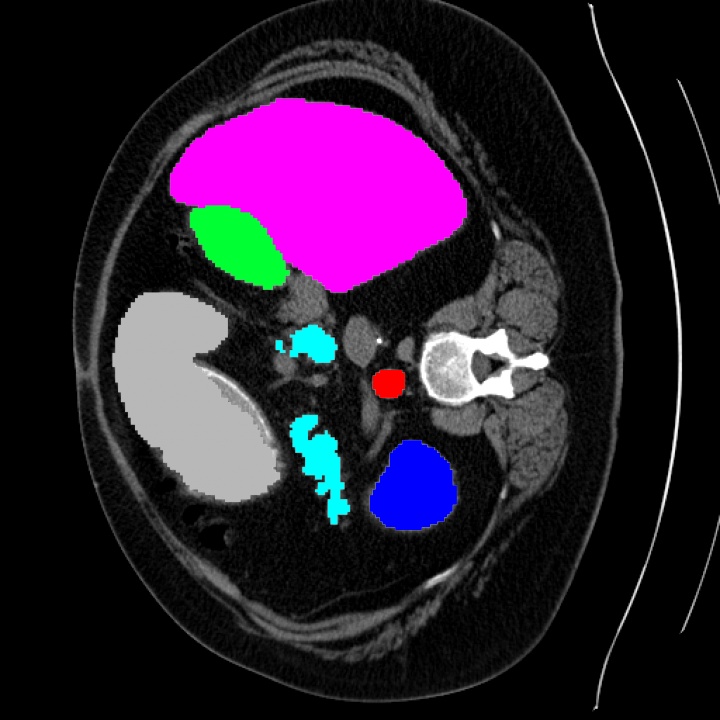}}\vspace{0.15cm} \\    
 \raisebox{-.5\height}{\includegraphics[width=3cm]{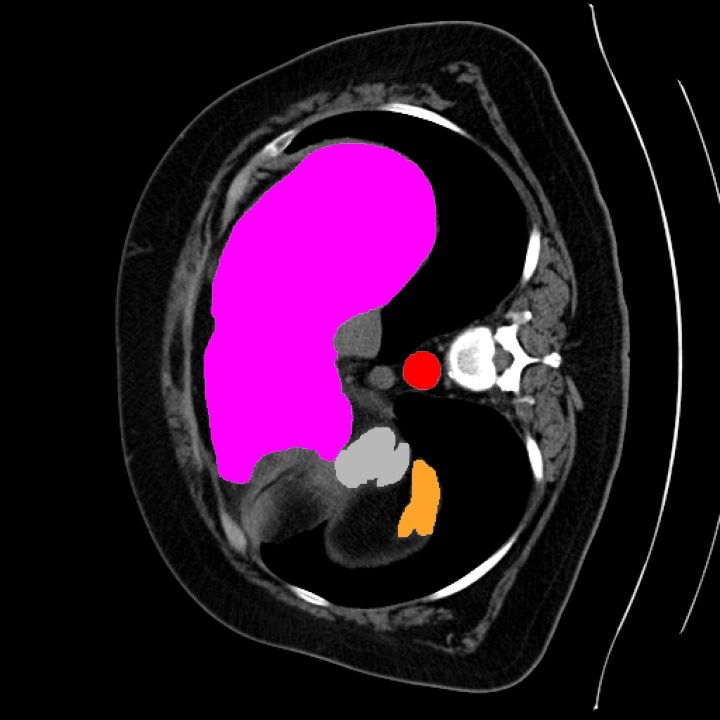}} &
 \raisebox{-.5\height}{\includegraphics[width=3cm]{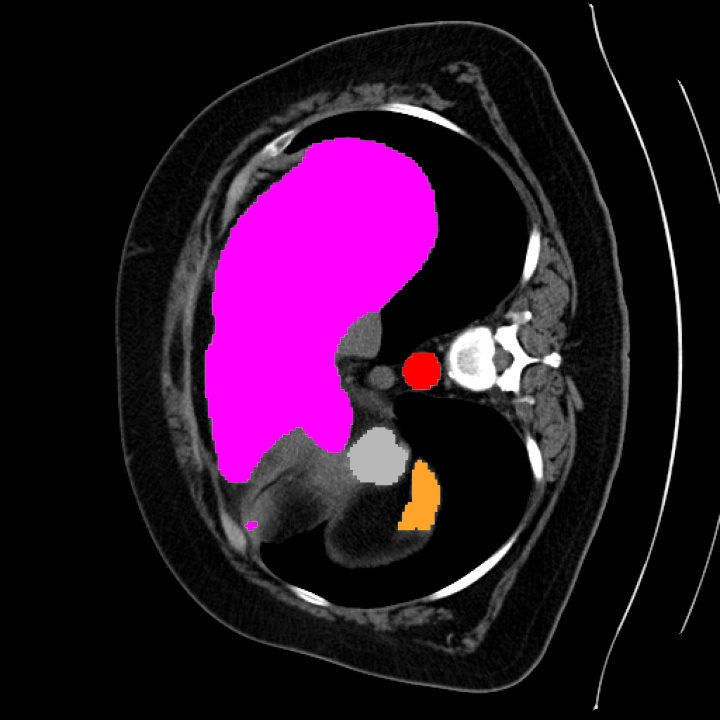}} &
    \raisebox{-.5\height}{\includegraphics[width=3cm]{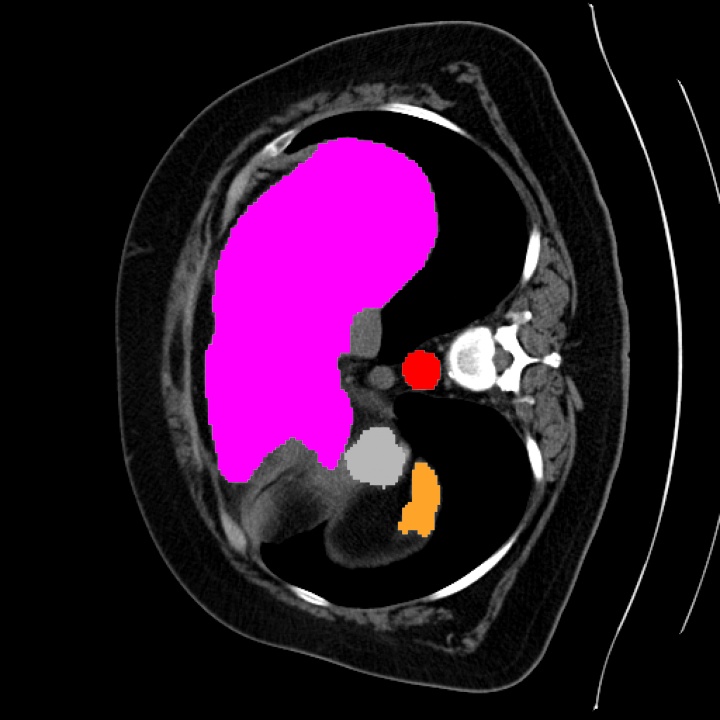}}\vspace{0.15cm} \\    
 \raisebox{-.5\height}{\includegraphics[width=3cm]{figures/comparison/unet/case0038_gt_68.jpg}} &
 \raisebox{-.5\height}{\includegraphics[width=3cm]{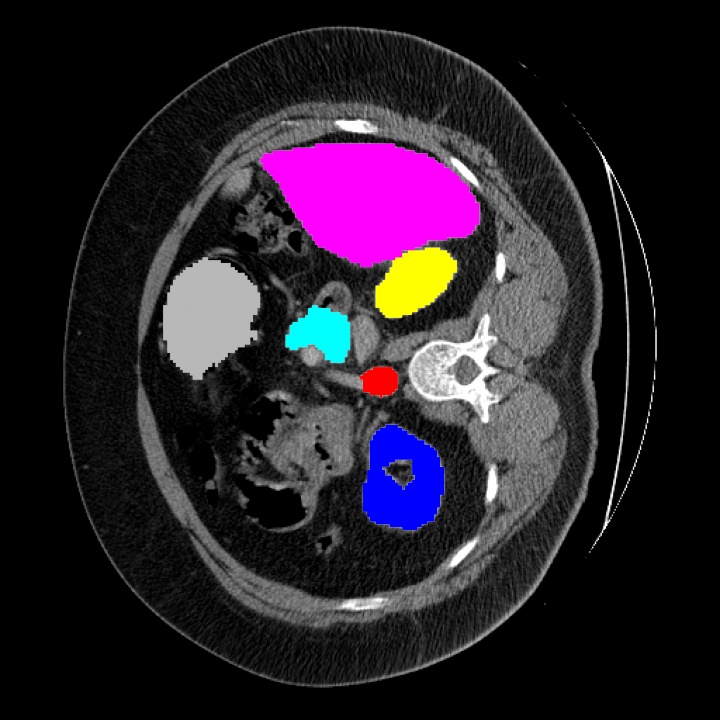}} &
    \raisebox{-.5\height}{\includegraphics[width=3cm]{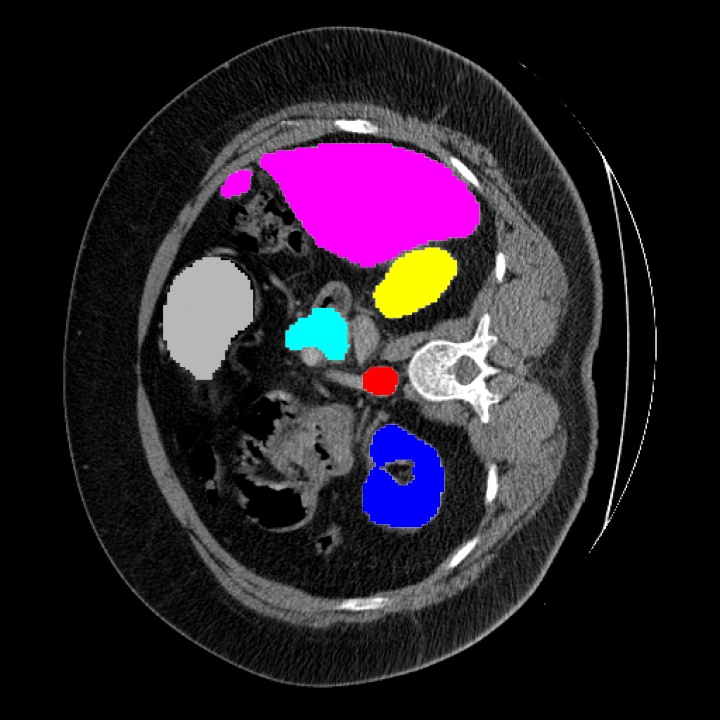}} \vspace{0.15cm}\\  
 \raisebox{-.5\height}{\includegraphics[width=3cm]{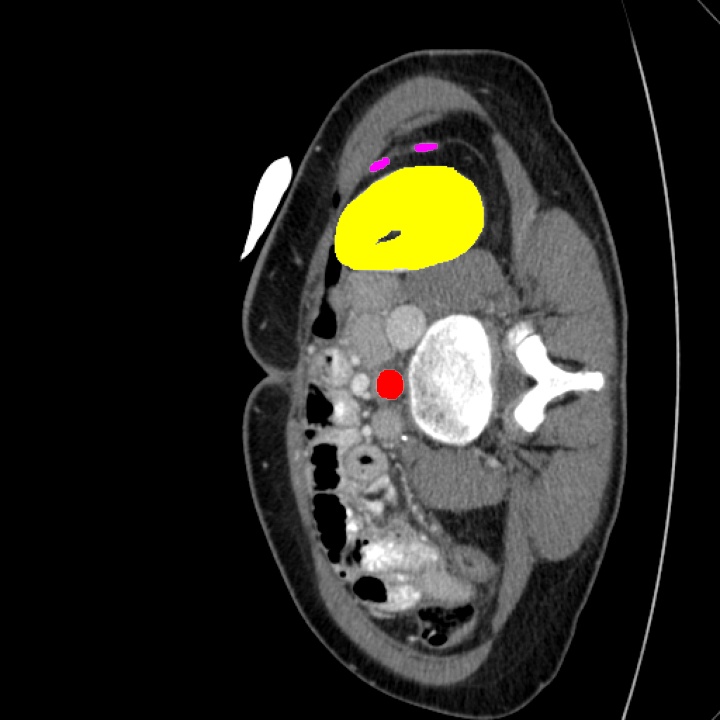}} &
 \raisebox{-.5\height}{\includegraphics[width=3cm]{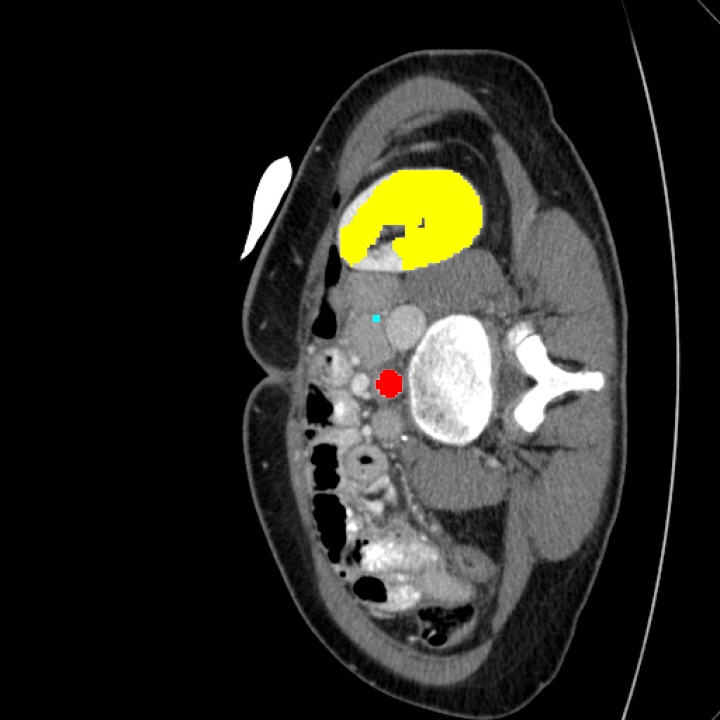}} &
    \raisebox{-.5\height}{\includegraphics[width=3cm]{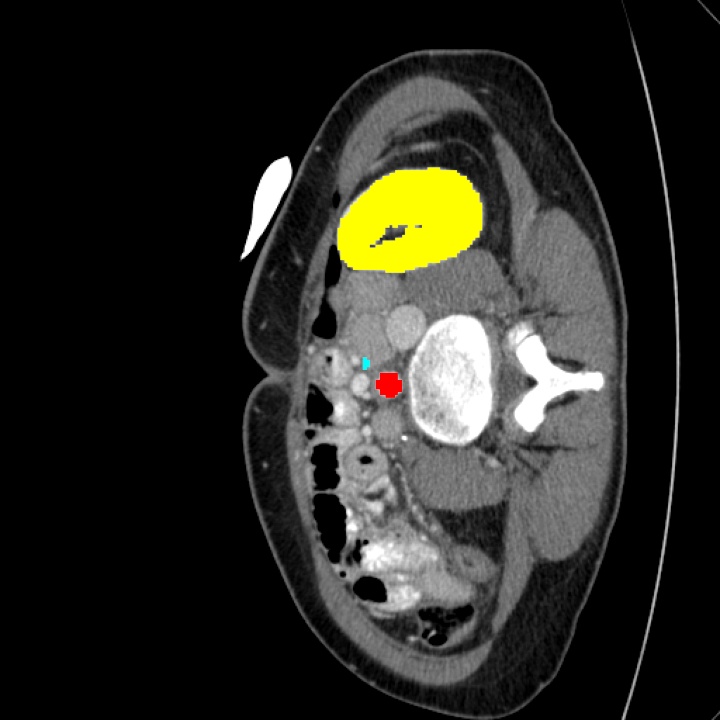}} \vspace{0.15cm}\\  
 \end{tabular}
}
    \caption{\textbf{Some qualitative results on comparison of \methodName{} against U-Net on the test data of the Synapse dataset \cite{synapsedataset}.} (\begingroup\setlength{\fboxsep}{0pt}\fcolorbox{black}{red}{\rule{0pt}{7pt}\rule{7pt}{0pt}}\endgroup~Aorta,
\begingroup\setlength{\fboxsep}{0pt}\fcolorbox{black}{synapse1}{\rule{0pt}{7pt}\rule{7pt}{0pt}}\endgroup~Gallbladder,
\begingroup\setlength{\fboxsep}{0pt}\fcolorbox{black}{blue}{\rule{0pt}{7pt}\rule{7pt}{0pt}}\endgroup~Left Kidney,
\begingroup\setlength{\fboxsep}{0pt}\fcolorbox{black}{yellow}{\rule{0pt}{7pt}\rule{7pt}{0pt}}\endgroup~Right Kidney,
\begingroup\setlength{\fboxsep}{0pt}\fcolorbox{black}{synapse4}{\rule{0pt}{7pt}\rule{7pt}{0pt}}\endgroup~Liver,
\begingroup\setlength{\fboxsep}{0pt}\fcolorbox{black}{synapse5}{\rule{0pt}{7pt}\rule{7pt}{0pt}}\endgroup~Spleen,
\begingroup\setlength{\fboxsep}{0pt}\fcolorbox{black}{synapse2}{\rule{0pt}{7pt}\rule{7pt}{0pt}}\endgroup~Pancreas, and
\begingroup\setlength{\fboxsep}{0pt}\fcolorbox{black}{synapse3}{\rule{0pt}{7pt}\rule{7pt}{0pt}}\endgroup~Stomach)}
    \label{ch:res:fig:synapse:figures:segperf} 
\end{figure}

We show the results of our experiments on the Synapse dataset \cite{synapsedataset} both quantitatively and qualitatively in \autoref{ch:res:synapse:finaltable} and \autoref{ch:res:fig:synapse:figures:segperf}. The average dice score and HD95 has improved respectively by $5.13$ percent and 7.76 points compared to Swin-Unet. The qualitative and quantitative results makes us believe that the performance differences among the compared methods vary based on the shape and the size of the segmented anatomy. It can be observed that for the larger organs, the results have less variations. For instance in \autoref{ch:res:fig:synapse:figures:segperf}, we see that the dice score in the liver has the highest value among abdomen organs, and despite the fact that our model outperformed others, the improvement is marginal and only around 1.05$\%$. We also observe that the methods are sensitive to irregularity in shape and compactness of the target anatomy. For instance, in the case of Aorta, our method, U-Net and Attn-Unet have similar performances in terms of dice score. Interestingly, transformer based-architectures have lower performance. The usual skip-connections in U-Net without any interference might be the reason. We observe similar behavior in \autoref{tab:ablation} as well. It can be seen that with lowering the effect of skip connections from the attention map branch in the last layer, the performance of \textit{LV} has improved, which is the only label \autoref{ch:res:table:acdc} for which U-Net is marginally better. \begin{table}[htb]
    \centering
    \caption{Comparison of our method against related work on the ACDC \cite{bernard2018deepACDC} dataset ($^*$ obtained from \cite{chen2021transunet}, and $^\dag$ trained by us).}
    \label{ch:res:table:acdc}
    \resizebox{\linewidth}{!}{%
    \begin{tabular}{l|c|c|c|c}
    \toprule
    \textbf{Method} & \textbf{RV} & \textbf{Myo} & \textbf{LV} & \textbf{DSC (\%)}\\ 
    \midrule
      R50 U-Net$^*$ \cite{alam2020automatic}             & 84.62 & 84.52 & 93.68 & 87.60 \\
      R50 AttnUNet$^*$ \cite{schlemper2019attention}    & 83.27 & 84.33 & 93.53 & 86.90 \\
      ViT-CUP$^*$ \cite{dosovitskiy2020imageVIT} 		                    & 80.93 & 78.12 & 91.17 & 83.41 \\
      R50 ViT$^*$ \cite{dosovitskiy2020imageVIT}                           & 82.51 & 83.01 & 93.05 & 86.19 \\
      TransUNet \cite{chen2021transunet}                & 86.67 & 87.27 & 95.18 & 89.71 \\
      Swin-Unet \cite{cao2021swinunet}                  & 85.77 & 84.42 & 94.03 & 88.07 \\
      MT-UNet \cite{wang2021mixed}                      & 86.64 & 89.04 & 95.62 & 90.43 \\ 
      U-Net$^\dag$ \cite{ronneberger2015u} & 89.67 & 89.27 & \textbf{95.76} & 91.57  \\  
      TransCASCADE \cite{rahman2023medical} & 89.14 & \textbf{90.25} & 95.50 & 91.63 \\ \midrule
      \methodName{} (Ours) & \textbf{90.53} & 89.52 & 95.63 & \textbf{91.90} \\ \bottomrule 
    \end{tabular}%
    }
\end{table}

\subsubsection{Ablation Study}
In \autoref{tab:ablation}, we present an ablation study of the proposed method. We analyze the role of different skip connections by gradually adding them from the first / last block to the bottleneck. 
\paragraph{Single Encoder} In this experiment, we present the results of the model with a single encoder, where the attention map visualizations are simply concatenated together with the input image to a single encoder. The basis for proposing a dual encoder network for our method is that the original image and the attention maps are from different domains and allowing their features to be extracted by two separate encoders provide richer features at the bottleneck. This can be seen in \autoref{tab:ablation}, as it shows that the dual encoder architecture achieves $1.5\%$ higher dice score compared to the single one.
\paragraph{Residual Connections} The results of the ablation study in \autoref{tab:ablation} show that having all four skip connections from the attention encoder to the decoder give the best overall performance; however, this is not the case for all of the regions. We observe that the removing the last skip connection has an advantage in \textit{LV}, and we speculate that this effect might be due to the fact that some shapes, maybe because of simplicity, benefit from using the priors that are directly obtained from the shapes and the priors from transformers makes the inference more complex; therefore, lowering the effect of the skip-connection in the last layer of the attention encoder yields better performance 

\subsubsection{Computational Cost}
We have compared the number of parameters that are used in transformer-based and state-of-the-art models in medical image segmentation in \autoref{ch:res:table:numparameters}, and as it can be observed, the number of parameters in some of the models is comparably large, and in turn, the training cost of the model is expected to be higher. On the other hand it is assumed that in most cases, the transformer block requires more training to capture the inductive bias as is reported by Dosovitskiy \cite{dosovitskiy2020imageVIT}. It can be therefore considered to also affects the training speed and cost even further. Nevertheless, we reported the number of multiple-accumulate operations(MAC) that can provide an idea about the inference time of the models. As can be seen, our main model with a lower amount of trainable parameters is able to achieve better performance compared to models like MT-UNet \cite{mtunetgithub} or TransUNet \cite{chen2021transunet}.
\begin{table}[htb]
    \centering
    \caption{Ablation study of our model on ACDC \cite{bernard2018deepACDC}}
    \label{tab:ablation}
    \resizebox{\linewidth}{!}{%
    \begin{tabular}{l|cccc|c|c|c|c}
    \toprule
    \multicolumn{1}{c|}{\multirow{3}{*}{\Centerstack{\textbf{Encoder}}}} & \multicolumn{4}{c|}{\multirow{2}{*}{\Centerstack{\textbf{Att. Skip-}\\\textbf{Connection}}}} & \multirow{3}{*}{\Centerstack{\textbf{RV}}}  & \multirow{3}{*}{\Centerstack{\textbf{Myo}}}    & \multirow{3}{*}{\Centerstack{\textbf{LV}}} & \multirow{3}{*}{\Centerstack{\textbf{DSC (\%)}}} \\
        \multicolumn{1}{c|}{} & \multicolumn{4}{c|}{} &  &  &  &   \\
    \multicolumn{1}{c|}{} & \circled{1} & \circled{2} & \circled{3} & \circled{4} &  &  &  &   \\ \midrule
    Single & - & - & - & - & 88.69 & 87.22 & 95.25 & 90.38 \\ \midrule 
     \multirow{4}{*}{Dual} & 0 & 0 & 0 & 0 & 89.83 & 89.29 & 95.67 & 91.60 \\ 
     & 0 & 0 & 0 & 1 & 90.23 &89.27 & 95.65 & 91.71 \\ 
     & 0 & 1 & 1 & 1 & 90.39 & 89.42  & \textbf{95.75} & 91.85 \\ 
     & 1 & 1 & 1 & 1 & \textbf{90.53} & \textbf{89.52} & 95.63 & \textbf{91.90} \\ \bottomrule 
    \end{tabular}%
    }
\end{table}

\begin{table}[!htbp]
\centering
\caption{Number of trainable parameters per model along with the amount of multiply–accumulate operations (MACs) necessary to process one image in some of the state-of-the-art image segmentation models and transformer-based architectures.}
\label{ch:res:table:numparameters}
\resizebox{\linewidth}{!}{%
\begin{tabular}{l|cc} 
\toprule
\multicolumn{1}{c|}{\multirow{2}{*}{\textbf{Method}}}  
& \multirow{2}{*}{\Centerstack{\textbf{Number of }\\\textbf{Trainable Parameters}}} & \multirow{2}{*}{\Centerstack{\textbf{MACs}}}\\
&& \\ \toprule 
nnUNet\cite{tang2019nodulenet} & 412.65M & 206.33G          \\
TransUNet \cite{chen2021transunet} & 96.07M &24.17G          \\
UNETR \cite{tang2021self}  & 92.58M & 20.60G          \\
MT-UNet \cite{wang2021mixed} &     75.07M &44.80G          \\ 
SwinUNETR\cite{lin2021ds} & 61.98M &197.42G          \\
Swin-Unet \cite{cao2021swinunet} & 41.34M & 8.73G \\
\midrule
\methodName{} (Ours) &      32.79M &43.95G            \\
\end{tabular}
}
\end{table}

\section{Limitations}
Our proposed pipeline receives pairs of image-attention map visualizations, which do not change during the training. The transformer block in our method can also be fine-tuned for different datasets for further improvement, but on the other hand, other models that incorporate the transformer blocks into their architecture are easier to be fine-tuned. If the generated attention-map visualizations capture irrelevant features, as seen in \autoref{ch:res:visualization}, especially visible in Attention Head 3 in sample B, it might give incorrect bias for the segmentation task.

Ideally, to demonstrate the full potential of these architectures, utilizing a ViT/DINO foundational model in 3D, pre-trained in a self-supervised manner on large amounts of unannotated medical images is preferred. However, such a model does not exist, and training our own backbone would require significant amounts of data and computational resources. Instead, we use this work as an opportunity to investigate how to optimally incorporate a 2D DINO backbone for medical image segmentation and showcase its effectiveness. In our results, we are able to show that our two simple 2D CNN architectures utilizing DINO attention maps are sufficient to outperform many reference and SOTA approaches, even those trained in 3D. In fact, our 2D method is also outperformed a 3D hybrid ViT-CNN architectures that is pre-trained on a large data \cite{hatamizadeh2022swin} on HD95 metric (ours: 13.79 compared to 20.53), whose attention maps are not directly usable in our approach as those from 2D DINO. This work, therefore, opens the path for new possibilities when novel backbone architectures become available.
\section{Conclusion}
\label{sec:conclusion}
In this work, we presented a simple yet effective model for semantic segmentation of medical images called \methodName{}. With that architecture, we showed that the self-attention map visualizations in transformers trained in a self-supervised manner, such as DINO could capture meaningful features that can be directly used as input for improving medical image segmentation. We used a model pretrained on a common computer vision dataset (here DINO), and we showed its effectiveness of the extracted features in medical image segmentation. Unlike other methods that incorporate the transformer blocks in the main architecture, our architecture does not depend on the transformer block in the run-time, and it can achieve state-of-the-art performance on two medical image segmentation benchmarks. We also presented an ablation study on different customization of the architecture. The proposed method can open the path for future architecture designs that aim to be lightweight yet take advantage of the representation power of transformers in their pipeline.

{\small
\bibliographystyle{ieee_fullname}
\bibliography{11_references}
}

\end{document}